\documentclass[10pt,twocolumn,letterpaper]{article}

\usepackage{cvpr}              

\usepackage{graphicx}
\usepackage{amsmath}
\usepackage{amssymb}
\usepackage{booktabs}

\usepackage[pagebackref,breaklinks,colorlinks]{hyperref}

\usepackage[capitalize]{cleveref}
\crefname{section}{Sec.}{Secs.}
\Crefname{section}{Section}{Sections}
\Crefname{table}{Table}{Tables}
\crefname{table}{Tab.}{Tabs.}

\usepackage{enumitem}    
\usepackage{efbox}       
\usepackage{xcolor}      
\usepackage[ruled,lined,linesnumbered,commentsnumbered]{algorithm2e}
\usepackage{multirow}
\usepackage{bbding}

\usepackage[accsupp]{axessibility}  

\def\cvprPaperID{2049} 
\def\confName{CVPR}
\def\confYear{2023}

\begin{document}

\title{Boost Vision Transformer with GPU-Friendly Sparsity and Quantization}

\author{Chong Yu\textsuperscript{1,2} \quad\quad Tao Chen\textsuperscript{3,*} \quad\quad Zhongxue Gan\textsuperscript{1,}\thanks{Tao Chen and Zhongxue Gan are corresponding authors.} \quad\quad Jiayuan Fan\textsuperscript{1}\\
\textsuperscript{1}Academy for Engineering and Technology, Fudan University
\quad \textsuperscript{2}NVIDIA Corporation
\\
\textsuperscript{3}School for Information Science and Technology, Fudan University
\\
{\tt\small 21110860050@m.fudan.edu.cn; \{eetchen, ganzhongxue, jyfan\}@fudan.edu.cn}
}
\maketitle

\begin{abstract}
   The transformer extends its success from the language to the vision domain. Because of the 
   stacked self-attention and cross-attention blocks
   , 
   the acceleration deployment of vision transformer on GPU hardware is challenging and also rarely studied. This paper thoroughly designs a compression scheme to maximally utilize the \textbf{GPU-friendly 2:4 fine-grained structured sparsity and quantization}. Specially, an original large model with dense weight parameters is first pruned into a sparse one by 2:4 structured pruning, which considers the GPU's acceleration of 2:4 structured sparse pattern with FP16 data type, then the floating-point sparse model is further quantized into a fixed-point one by sparse-distillation-aware quantization aware training, which considers GPU can provide an extra speedup of 2:4 sparse calculation with integer tensors. A mixed-strategy knowledge distillation is used during the pruning and quantization process. The proposed compression scheme is flexible to support supervised and unsupervised learning styles. Experiment results show \textbf{GPUSQ-ViT} scheme achieves state-of-the-art compression by reducing vision transformer models \textbf{6.4-12.7$\times$} on model size and \textbf{30.3-62$\times$} on FLOPs with negligible accuracy degradation on ImageNet classification, COCO detection and ADE20K segmentation benchmarking tasks. Moreover, \textbf{GPUSQ-ViT} can boost actual deployment performance by \textbf{1.39-1.79$\times$} and \textbf{3.22-3.43$\times$} of latency and throughput on A100 GPU, and \textbf{1.57-1.69$\times$} and \textbf{2.11-2.51$\times$} improvement of latency and throughput on AGX Orin.
\end{abstract}

\section{Introduction}
\label{sec:introduction}

Transformer-based neural models~\cite{sutskever2014sequence} have garnered immense interest recently due to their effectiveness and generalization across various applications. Equipped with the attention mechanism~\cite{vaswani2017attention} as the core of its architecture, transformer-based models specialize in handling long-range dependencies, which are also good at extracting non-local features~\cite{dosovitskiy2020image}~\cite{carion2020end} in the computer vision domain. With comparable and even superior accuracy than the traditional convolution neural networks (CNN)~\cite{he2016deep}~\cite{tan2019efficientnet}, more vision transformer models are invented and gradually replace the CNN with state-of-the-art performance on image classification~\cite{liu2021swin}~\cite{liu2022swin}, object detection~\cite{zhu2020deformable}~\cite{xu2021end}, and segmentation~\cite{xie2021segformer}~\cite{zhou2022understanding} tasks. Due to the vision transformer models having a generally weaker local visual inductive bias~\cite{dosovitskiy2020image} inherent in CNN counterparts, many transformer blocks are stacked for compensation. Moreover, the attention module in the transformer block contains several matrix-to-matrix calculations between key, query, and value parts~\cite{vaswani2017attention}. Such designs give the naive vision transformers more parameters and higher memory and computational resource requirements, causing high latency and energy consuming during the inference stage. \textit{It is challenging for actual acceleration deployment in GPU hardware}.


Model compression techniques to transfer the large-scale vision transformer models to a lightweight version can bring benefits to more efficient computation with less on-device memory and energy consumption. There are some previous studies to inherit CNN 
compression methods, including pruning~\cite{rao2021dynamicvit}~\cite{hou2022multi}, quantization~\cite{liu2021post}~\cite{li2022q}, 
distillation~\cite{yang2022vitkd}, and architecture search~\cite{chavan2022vision} on vision transformers. However, there are some drawbacks in 
previous studies:
\begin{itemize}[topsep=1pt,partopsep=0pt,itemsep=1.5pt,parsep=\parskip]
\item Most of these common methods aim to reduce the theoretical model size and Floating Point Operations (FLOPs). But it has been proved~\cite{mishra2021accelerating}~\cite{nvidiatensorrt} that smaller model sizes and FLOPs are not directly proportional to better efficiency on deployed hardware.
\item The compression patterns do not match hardware characteristics. For example, pruned~\cite{rao2021dynamicvit} or searched~\cite{chavan2022vision} vision transformer models have the unstructured sparse pattern in weight parameters, i.e., the distribution of non-zero elements is random. So deployed hardware can not provide actual speedup due to lacking the characteristics support for unstructured sparsity~\cite{nvidiaa100}.
\item How to keep the accuracy 
to the best with multiple compression methods and how to generalize on multiple vision tasks lack systematical investigation.
\end{itemize}

\begin{figure}[!htb]
\begin{center}
  \centering
  \includegraphics[width = 0.93\linewidth]{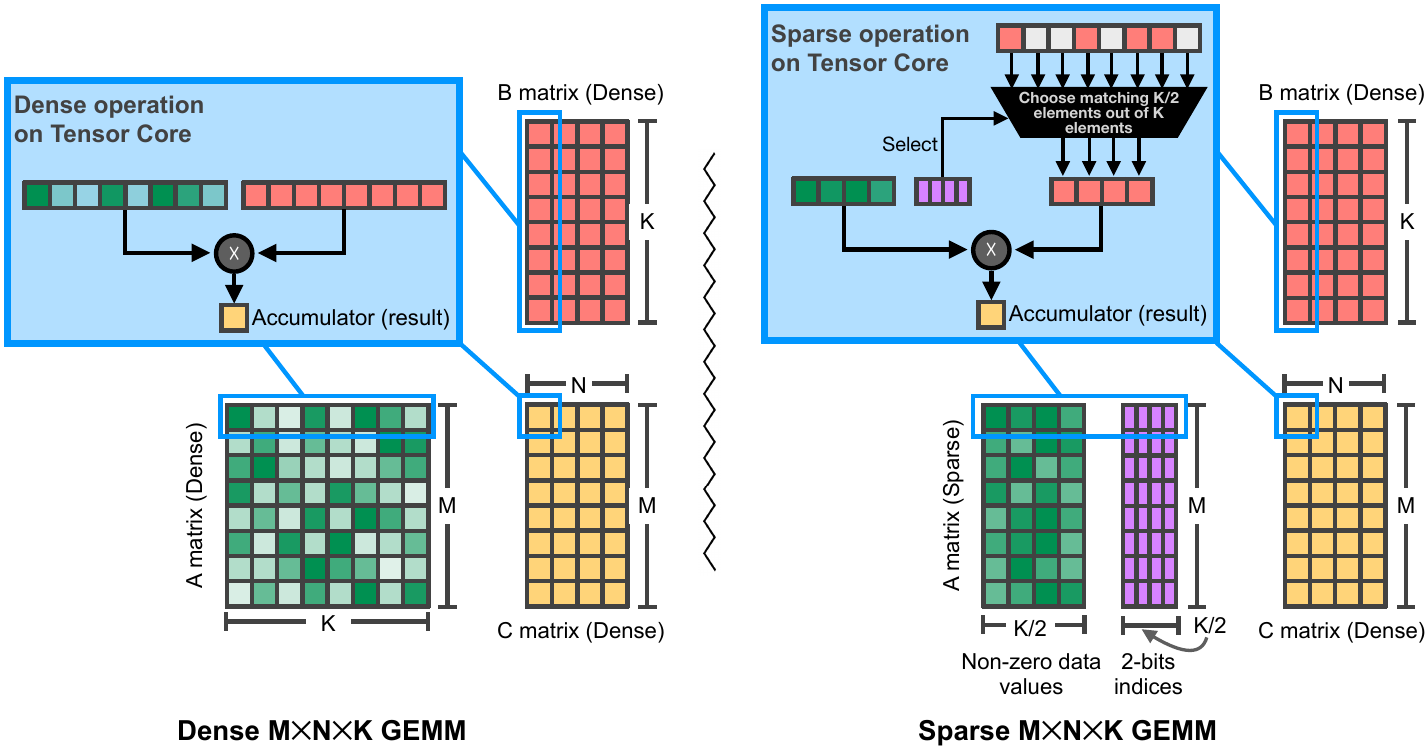}
\end{center}
\vskip -0.2in
	\caption{Comparison of computing a $M\times N\times K$ GEMM onto a Tensor Core.
    Dense matrix A of size $M\times K$ in \textit{\textbf{left side}} becomes $M\times\frac{K}{2}$ in \textit{\textbf{right side}} after compressing with \textit{\textbf{2:4 fine-grained structured sparse pattern}}. Sparse Tensor Core automatically picks only the elements from B according to the nonzero elements in A. Comparing the dense and sparse GEMMs, B and C are the same dense $K\times N$ and $M\times N$ matrices, respectively. By skipping the unnecessary multiplications of redundant zeros, sparse GEMM accelerate the dense GEMM with $2\times$.} 
	\label{Fig.sparse_tensor_core}
\vskip -0.10in
\end{figure}

General Matrix Multiplication (GEMM) is the fundamental implementation inside the common parts of vision transformers, such as convolution, linear projection, and transformer blocks. A specific acceleration unit called Tensor Core~\cite{nvidiatc} is firstly introduced in NVIDIA Volta GPU~\cite{nvidiav100} to accelerate these GEMM instructions and further enhanced to support sparse GEMM in 
Ampere GPU~\cite{nvidiaa100}. To make the GPU hardware efficient for sparse GEMM, a constraint named \textit{\textbf{2:4 fine-grained structured sparsity}}~\cite{mishra2021accelerating} is imposed on the allowed sparsity pattern, i.e., two values from every four contiguous elements on rows must be zero. Due to the 2:4 sparsity support on GPU Tensor Core hardware, sparse GEMM can reduce memory storage and bandwidth by almost $2\times$ and provide $2\times$ math throughput compared to dense GEMM by skipping the redundant zero-value computation, as shown in Figure~\ref{Fig.sparse_tensor_core}. Ampere GPU supports various numeric precision for 2:4 sparsity, including FP32, FP16, INT8, and INT4, etc.

Inspired by GPU's acceleration characteristic for 2:4 fine-grained structured sparse pattern with various low-precision operators, we thoroughly design the compression scheme \textbf{GPUSQ-ViT} by utilizing the \textbf{GPU}-friendly \textbf{S}parsity and \textbf{Q}uantization to boost deployment efficacy for \textbf{Vi}sion \textbf{T}ransformer models, especially on GPU platforms. \textbf{GPUSQ-ViT} contains two main workflows. Firstly, 2:4 sparse pruning with knowledge distillation~\cite{hinton2015distilling} (KD) is proposed to compress the specific structures in vision transformer architecture, e.g., transformer block, patch embedding, to be GPU-friendly. Secondly, we further quantize the sparse model through sparse-distillation-aware Quantization Aware Training~\cite{mckinstry2018discovering} (QAT). To measure the influence of quantization errors, we use the feature-based distillation loss in the sparse pruning workflow as the weight factor. The feature-based KD utilizes the scale factor in the quantization compression workflow, which can best compensate for the final compressed model's accuracy. We demonstrate that 
\textbf{GPUSQ-ViT} can generally apply to 
vision transformer models and benchmarking tasks, with state-of-the-art theoretical metrics on model size and FLOPs. Moreover, as \textbf{GPUSQ-ViT} compresses with GPU-friendly patterns, the compressed models can achieve state-of-the-art deployment efficacy on GPU platforms. Our main contributions include:
\begin{itemize}[topsep=1pt,partopsep=0pt,itemsep=1.5pt,parsep=\parskip]
\item Unlike previous compression methods only aiming at reducing theoretical metrics, we propose \textbf{GPUSQ-ViT} from the perspective of GPU-friendly 2:4 sparse pattern with low-precision quantization for the first time, achieving 
GPU acceleration of 4 times than prior arts. 
\item \textbf{GPUSQ-ViT} combines feature-based KD with sparse pruning and QAT, which can best compensate for sparse and quantized models' accuracy. 
\item \textbf{GPUSQ-ViT} can apply to various vision transformer models and benchmarking tasks, with proven state-of-the-art efficacy on model size, FLOPs, and actual deployment performance on multiple GPUs. Moreover, \textbf{GPUSQ-ViT} can work without ground truth label annotations in an unsupervised learning style.
\end{itemize}

\section{Related work}
\label{sec:related_work}

\subsection{Sparsity in model compression}
\label{sub_sec:related_work_sparsity}
Sparsity is a typical pattern~\cite{han2015learning} in the deep learning paradigm, which can help to save the computational power as well as reduce the memory bandwidth and storage burden~\cite{mishra2021accelerating}. 
Sparsity has different granularities~\cite{mao2017exploring}, e.g., we can generate the filter-level, kernel-level, vector-level, and element-level sparsity~\cite{mao2017exploring} in a weight tensor from coarse to fine granularity. The coarse-grained sparsity has a regular sparse pattern which can facilitate acceleration with algebra libraries~\cite{mishra2021accelerating}. The fine-grained sparsity leads to a more irregular sparse pattern which is not friendly for acceleration, but it can achieve a higher sparse ratio without harming model accuracy~\cite{yang2020winograd}~\cite{yu2020self}. Many previous efforts~\cite{cai2019once}~\cite{yu2020self}~\cite{li2020gan} have explored the sparse granularity to balance accuracy influence with real performance benefits.

Several efforts explored to compress the vision transformers with sparsity. Inspired by the phenomenon that the vision transformers take effect only according to a subset of most informative tokens~\cite{rao2021dynamicvit}, we can generate the sparse tokens by pruning the less informative ones. The redundant tokens are pruned based on the inputs, spatial attention mechanism~\cite{ryoo2021tokenlearner}, or multi-head interpreter~\cite{pan2021ia} in a dynamical~\cite{rao2021dynamicvit} or patch-slimming manner~\cite{tang2022patch}.

Other efforts are explored on how to prune the components inside the basic structure in vision transformers, i.e., the multi-head attention block (MHA)~\cite{vaswani2017attention}. For example, a successful trial~\cite{zhu2021vision} is first to learn the importance of each component in MHA 
by training with sparse regularization, then pruning the less important ones to obtain the sparse 
MHA. Other strategies aim to sparsify the attention heads and reduce the sequence length in an MHA structure based on specific numerical metrics~\cite{wang2021spatten} or searched optimal policy~\cite{hou2022multi}. A more aggressive approach is pruning the entire MHA blocks to generate a sparse Mixture-of-Experts~\cite{hwang2022tutel} vision transformer or an extremely compact version~\cite{zhang2022minivit}. \textit{Most of the prior arts use model sizes and FLOPs as compression targets without considering the characteristics of deployed hardware.} We find low efficiency when deploying these compressed models on GPUs, which inspires us to \textbf{\textit{design the compression scheme with a GPU-friendly sparse pattern}}. Based on prior arts, weight multiplexing~\cite{zhang2022minivit} or knowledge distillation~\cite{yu2022unified}~\cite{yang2022vitkd} are effective to compensate for the accuracy loss.

\subsection{Quantization in model compression}
\label{sub_sec:related_work_quantization}
Quantization is another orthogonal technique in the model compression area. It refers to the technique~\cite{wu2020integer} of applying alternative formats other than the standard 32-bit single-precision floating-point (FP32) data type for weight parameters, inputs, and activations when executing a neural model. 
Quantization can significantly speed up the model inference performance because the low-precision formats have higher computational throughput support in many processors~\cite{nvidiaa100}~\cite{jouppi2017datacenter}~\cite{arafa2019cascade}. Meanwhile, 
low-precision representation helps to reduce the memory bandwidth pressure and can save much memory-system operation time with the cache utilization improvement.

Post Training Quantization (PTQ)~\cite{krishnamoorthi2018quantizing} and Quantization Aware Training (QAT)~\cite{mckinstry2018discovering} are two main strategies in quantization. PTQ directly calibrates on limited sample inputs~\cite{migacz2017nvidia} to find the optimal clipping threshold and the scale factor to minimize the quantization noise~\cite{banner2019post}. PTQ is preferred~\cite{shomron2021post} when without access to the whole training dataset~\cite{li2022patch}. However, it is a non-trivial effort~\cite{liu2021post}~\cite{yuan2021ptq4vit}~\cite{lin2022fq}~\cite{li2022q} to ensure the PTQ quantized vision transformer model without an apparent accuracy decrease. And the accuracy degradation is more serious when going below 8 bits formats~\cite{shomron2021post}. QAT inserts the quantization and de-quantization nodes~\cite{nvidiatensorrt} into the float-point model structure, then undergo the fine-tuning process to learn the scale factor adjustment with minimal influence on accuracy~\cite{mckinstry2018discovering}. Considering some activation structures like GeLU~\cite{hendrycks2016gaussian} and Swish~\cite{ramachandran2017searching} are more sensitive~\cite{li2022q} than ReLU~\cite{agarap2018deep}, some efforts are made to design the specific QAT~\cite{li2022q}~\cite{li2022auto} for the vision transformers. Moreover, 
QAT can provide more quantization robustness for lower-bit formats~\cite{li2022q}.

Previous efforts to design the PTQ and QAT approaches for vision transformer mainly focused on the accuracy improvement. \textbf{\textit{Due to the lack of hardware characters and acceleration library support}}, some quantized models using 6 bits~\cite{liu2021post} or float-point learnable bit-width like 3.7 bits~\cite{li2022q} to represent weights and activations\textbf{ \textit{cannot get the expected speedup on general acceleration hardware}}, like GPU~\cite{nvidiav100}~\cite{nvidiaa100} and TPU~\cite{sato2017depth}. Moreover, \textit{supporting the specific bit-width quantization}, like 6 bits, \textit{is a non-trivial effort}. End-users need to program the FPGA hardware~\cite{li2022auto} and develop specific bit-width libraries like Basic Linear Algebra Subprograms (BLAS)~\cite{li2019cpu}, which is a heavy burden for actual deployment.

\section{Boost vision transformer on GPU}
\label{sec:boost_vision_transformer_on_gpu}

\textbf{GPUSQ-ViT} mainly contains \textbf{2:4 structured sparse pruning} and \textbf{sparse-distillation-aware QAT} workflows. We further explain the 2:4 sparse pattern in section~\ref{sub_sec:2_4_sparsity_on_gpu}, and how to compress each part of a vision transformer model according to the 
2:4 sparse pattern in sections~\ref{sub_sec:apply_sparsity_in_transformer_block} and~\ref{sub_sec:apply_sparsity_in_patch_embedding}. Section~\ref{sub_sec:overall_compression_method} describes the \textbf{GPUSQ-ViT} design as a whole.

\subsection{Fine-grained structured sparsity on GPU}
\label{sub_sec:2_4_sparsity_on_gpu}
As shown in Figure~\ref{Fig.sparse_tensor_core}, the sparse GEMM performs the \textit{sparse matrix $\times$ dense matrix = dense matrix} operation by skipping the redundant zero-value computation with sparse Tensor Core acceleration. For example, matrix A of size $M\times K$ follows the \textit{\textbf{2:4 fine-grained structured sparse pattern}}, and the dense matrix B is of size $K\times N$. If we use the dense GEMM to calculate between matrices A and B, the zero values in A would not be skipped during computation. The entire $M\times N \times K$ dense GEMM will calculate the result matrix C with $M\times N$ size in \textit{T} GPU cycles. If we use the sparse GEMM, only the non-zero elements in each row of matrix A and the corresponding elements from matrix B, which sparse Tensor Core automatically picks out without run-time overhead, are calculated. So the entire $M\times N \times K$ sparse GEMM will also calculate the same result matrix C with $M\times N$ size but only needs \textit{T/2} GPU cycles, leading to $2\times$ math throughput speedup.

\fboxsep = 1.5pt    
\fboxrule = 1.0pt   
\begin{figure}[!htb]
\vskip -0.10in
\begin{center}
  \centering
  \fcolorbox{teal}{white}{\includegraphics[width = 0.97\linewidth]{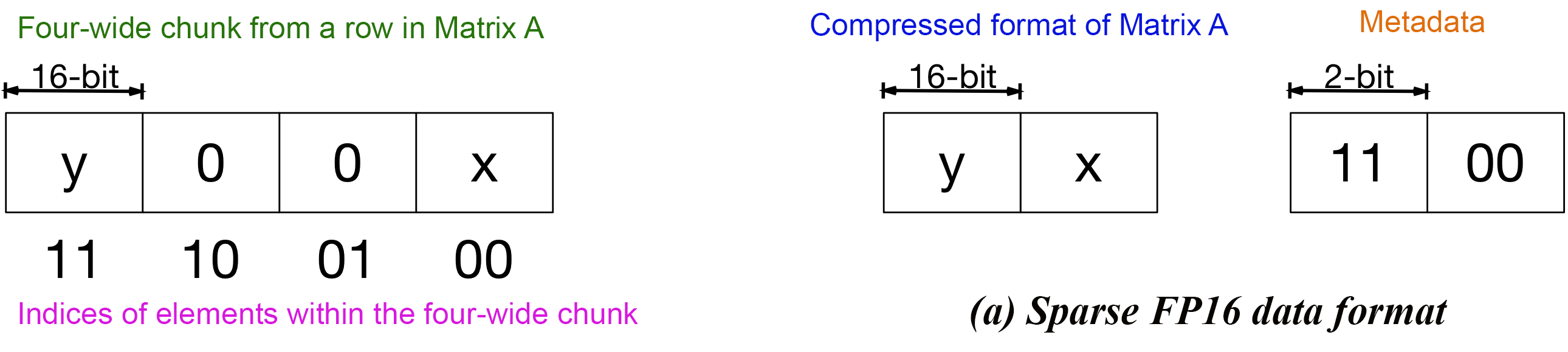}}
  \fcolorbox{blue}{white}{\includegraphics[width = 0.97\linewidth]{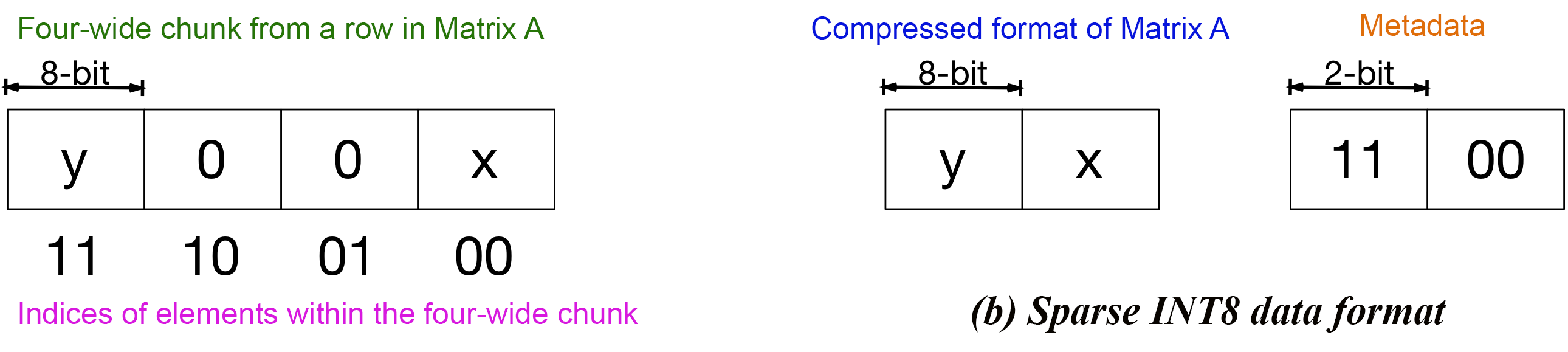}}
  \fcolorbox{orange}{white}{\includegraphics[width = 0.97\linewidth]{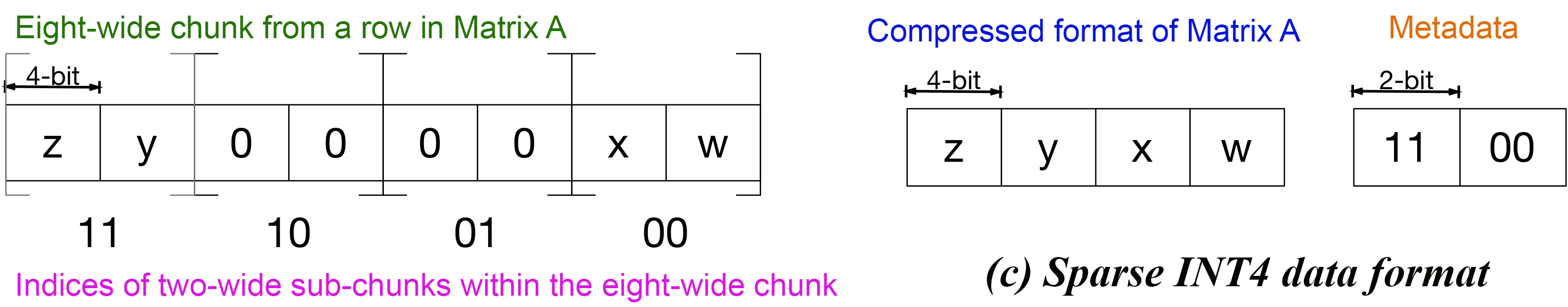}}
\end{center}
\vskip -0.2in
	\caption{Storage formats for \textit{\textbf{2:4 fine-grained structured sparse pattern}} and 
	metadata with FP16, INT8 and INT4 operators. (w,x,y,z denote the non-zero elements.)} 
	\label{Fig.sparse_metadata}
\vskip -0.15in
\end{figure}

The 2:4 sparsity uses 2-bit metadata per non-zero element to indicate the position of two non-zero elements in every four adjacent elements in a row of matrix A
with FP16 and INT8 data formats.
The 2:4 sparsity instruction for the INT4 data format differs from FP16 and INT8. Matrix A is defined as a pair-wise structured sparse at a granularity of 4:8. In other words, each chunk of eight adjacent elements in a row of matrix A has four zero and four non-zero values. Further, the zero and non-zero values are clustered in sub-chunks of two elements each within the eight-wide chunk, i.e., each two-wide sub-chunk within the eight-wide chunk must be all zeroes or all non-zeroes. Only the four non-zero values are stored in the compressed matrix, and two 2-bit indices in the metadata indicate the position of the two two-wide sub-chunks with non-zero values in the eight-wide chunk of a row of matrix A. 
In conclusion, the sparse format for FP16, INT8, and INT4 lead to 43.75\%, 37.5\%, and 37.5\% savings in storage. \textbf{GPUSQ-ViT} will firstly compress 
model as 2:4 FP16 sparse, then further quantize to 2:4 INT8 or INT4 sparse for best deployment efficiency.

Because the \textit{\textbf{2:4 fine-grained structured sparse pattern}} is well supported on NVIDIA GPU and corresponding libraries for math acceleration and memory saving, so we are motivated to \textit{\textbf{design the compression strategy for vision transformer models to meet such sparse pattern}}. Moreover, the 2:4 sparse GEMM supports low-precision formats like INT8 and INT4. So it is natural to \textit{\textbf{combine the sparsity and quantization in the proposed strategy jointly}} and further boost the actual deployment performance on GPUs.

\subsection{Apply structured sparsity in transformer block}
\label{sub_sec:apply_sparsity_in_transformer_block}

\begin{figure}[!htb]
\vskip -0.10in
\begin{center}
  \centering
  \includegraphics[width = 0.99\linewidth]{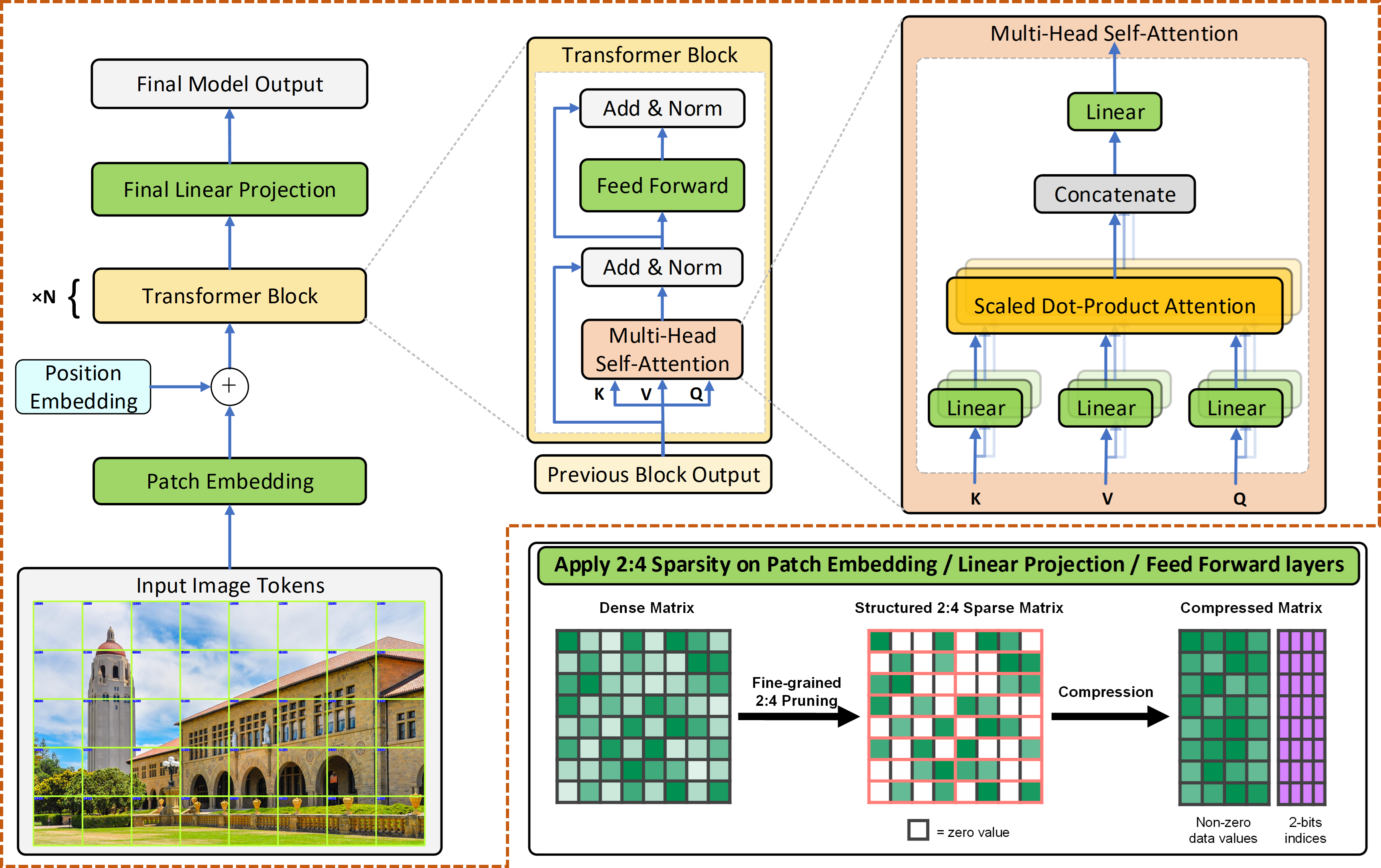}
\end{center}
\vskip -0.2in
	\caption{Illustration about applying the \textit{\textbf{2:4 fine-grained structured sparsity}} in vision transformer. The target layers include the patch embedding, final linear projection, as well as the feed forward and linear projection inside each transformer block.} 
	\label{Fig.apply_sparsity_for_vision_transformer}
\vskip -0.15in
\end{figure}

The transformer block~\cite{vaswani2017attention} is the fundamental building structure in various vision transformers. The majority of the weight parameters and the execution time are taken in stacked transformer blocks. For example, about 96\% of the weight parameters and 95\% of the inference time are from the transformer blocks in Swin Transformer~\cite{liu2021swin}. So we focus on how to apply the \textit{\textbf{2:4 fine-grained structured sparsity}} in the transformer block.

Transformer blocks used in vision transformer models are directly borrowed from~\cite{dosovitskiy2020image}~\cite{touvron2021training} or made tiny changes~\cite{liu2021swin}~\cite{wang2021pyramid} on the standard transformer block introduced in the naive attention mechanism~\cite{vaswani2017attention}. For example, the transformer block in the Swin Transformer model is built by replacing the standard multi-head attention module with a shifted windows attention module~\cite{liu2021swin}, with other layers kept the same as the standard transformer block. Without losing the generalization of the proposed method, we explore the utilization of 2:4 sparsity on a standard transformer block. \textit{\textbf{2:4 fine-grained structured sparsity}} accelerates GEMM operations, so the Q, K, and V projection layers, the linear projection layer in the multi-head attention module, and the linear projection layers in the feed-forward module are the proper targets to apply, as shown in the zoomed-in parts in Figure~\ref{Fig.apply_sparsity_for_vision_transformer}.

\subsection{Apply structured sparsity in patch embedding}
\label{sub_sec:apply_sparsity_in_patch_embedding}
The vision transformer paradigm splits each input image into small square patches~\cite{dosovitskiy2020image}
, and each image patch is treated as a token in the same way in the NLP domain. In vision transformer models, the following trainable linear embedding process is handled by a patch embedding layer and is usually implemented as a strided-convolution~\cite{dosovitskiy2020image}~\cite{liu2021swin}. Considering the input images are organized as an $N\times C\times H\times W$ batched data format, and each image will be divided into small patches with $P\times P$ square shape, where $N$ refers to batch size, $C$ refers to the number of the input channel, $H$ and $W$ refers to the height and width of an input image, $P$ refers to the size of each patch. So there will be $C\times(H\times W) / (P\times P)$ patches for each image, and each patch will be flattened as a token with shape $1\times P^2$. Suppose the given embedding dimension is denoted as $D_{embed}$. In that case, the patch embedding layer can be implemented with a convolution layer with $C$ as the input channel, $D_{embed}$ as the output channel, and kernel size and stride step equal to $P$. The total Floating Point Operations (FLOPs) of the patch embedding layer is $2\times N\times C\times H\times W\times D_{embed}$.

The strided-convolution layer is executed as an implicit GEMM~\cite{chellapilla2006high}~\cite{nvidiacutlass} on GPUs, which the \textit{\textbf{2:4 fine-grained structured sparsity}} can also accelerate, as shown in left-most of Figure~\ref{Fig.apply_sparsity_for_vision_transformer}. The implicit GEMM transfers the weight matrix of strided-convolution with $C\times P\times P$ as the width of matrix A, which is the target dimension to apply the 2:4 sparsity. It helps to save half of the total FLOPs
.

\subsection{Overall GPUSQ-ViT compression method}
\label{sub_sec:overall_compression_method}

\begin{figure*}[htb]
\vskip -0.10in
\begin{center}
  \centering
  \includegraphics[width = 0.95\linewidth]{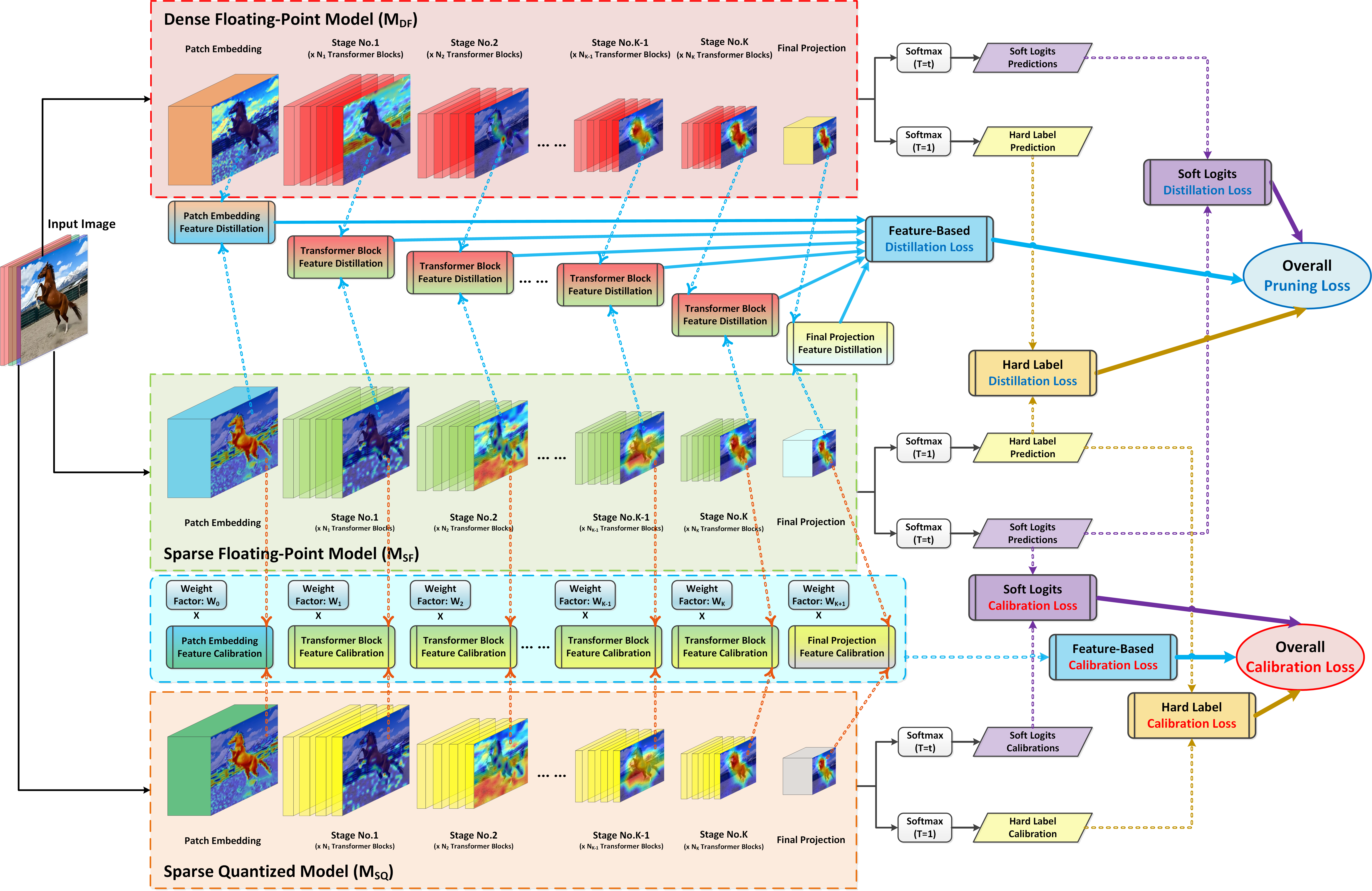}
\end{center}
\vskip -0.2in
	\caption{\textbf{GPUSQ-ViT} scheme with two sub-workflows
	. For the \textbf{2:4 structured sparse pruning workflow}, the dense floating-point model $\textbf{\emph{M}}_{DF}$ is compressed as the sparse floating-point model $\textbf{\emph{M}}_{SF}$. 
	Hard label, soft logits and feature-based distillation losses are accumulated as the overall sparse pruning loss. The sparse floating-point model $\textbf{\emph{M}}_{SF}$ is quantized as the sparse quantized model $\textbf{\emph{M}}_{SQ}$ for the \textbf{sparse-distillation-aware QAT workflow}. Hard label and soft logits calibration losses are obtained in a similar manner. Each feature maps calibration result is multiplied with a weight factor to indicate this layer's probability of having a real influence on 
	$\textbf{\emph{M}}_{SQ}$ model's final accuracy. 
	Three calibration losses are accumulated as the overall quantization calibration loss.}
	\label{Fig.overall_compression}
\vskip -0.2in
\end{figure*}

\textbf{GPUSQ-ViT} mainly contains \textbf{2:4 structured sparse pruning} and \textbf{sparse-distillation-aware QAT} workflows, as shown in Figure~\ref{Fig.overall_compression}. KD is applied in each workflow as auxiliary accuracy compensation.

\setlength{\parskip}{0.6em}
\noindent\textbf{2:4 Structured Sparse Pruning} aims to compress the dense floating-point model $\textbf{\emph{M}}_{DF}$ as the sparse floating-point model $\textbf{\emph{M}}_{SF}$. Based on 
Sections~\ref{sub_sec:apply_sparsity_in_transformer_block} and~\ref{sub_sec:apply_sparsity_in_patch_embedding}, we can compress each part of a vision transformer model according to the GPU-friendly 2:4 fine-grained structured sparse pattern. To best compensate for the accuracy of $\textbf{\emph{M}}_{SF}$, we apply KD~\cite{hinton2015distilling} which can effectively transfer the predicted hard label or soft logits from a teacher model with appealing performance to a student model
. If the student model wants to learn more, feature-based KD is applied to mimic the teacher model's feature maps. In 2:4 structured sparse pruning workflow, three KD strategies are jointly used.
\setlength{\parskip}{0em}

\fboxsep = 0.0pt    
\fboxrule = 1.0pt   
\begin{figure}[!htb]
\begin{center}
    \centering
    \begin{minipage}[b]{0.09\linewidth}
		\centering
		\textbf{\texttt{\fontsize{7.5pt}{\baselineskip}\selectfont No.}}
	\end{minipage}
    \begin{minipage}[b]{0.170\linewidth}
		\centering
		\textbf{\texttt{\fontsize{7.5pt}{\baselineskip}\selectfont Input}}
	\end{minipage}
	\begin{minipage}[b]{0.170\linewidth}
		\centering
		\textbf{\texttt{\fontsize{7.5pt}{\baselineskip}\selectfont Stage 1}}
	\end{minipage}
	\begin{minipage}[b]{0.170\linewidth}
		\centering
		\textbf{\texttt{\fontsize{7.5pt}{\baselineskip}\selectfont Stage 2}}
	\end{minipage}
	\begin{minipage}[b]{0.170\linewidth}
		\centering
		\textbf{\texttt{\fontsize{7.5pt}{\baselineskip}\selectfont Stage 3}}
	\end{minipage}
	\begin{minipage}[b]{0.170\linewidth}
		\centering
		\textbf{\texttt{\fontsize{7.5pt}{\baselineskip}\selectfont Stage 4}}
	\end{minipage}
\vskip 0.04in

    \begin{minipage}[b]{0.09\linewidth}
		\centering
		\textbf{\texttt{\fontsize{7.5pt}{\baselineskip}\selectfont (a-1)}}
	\end{minipage}
	\begin{minipage}[b]{0.170\linewidth}
		\centering
        \fcolorbox{green}{yellow}{\includegraphics[width = 0.99\linewidth]{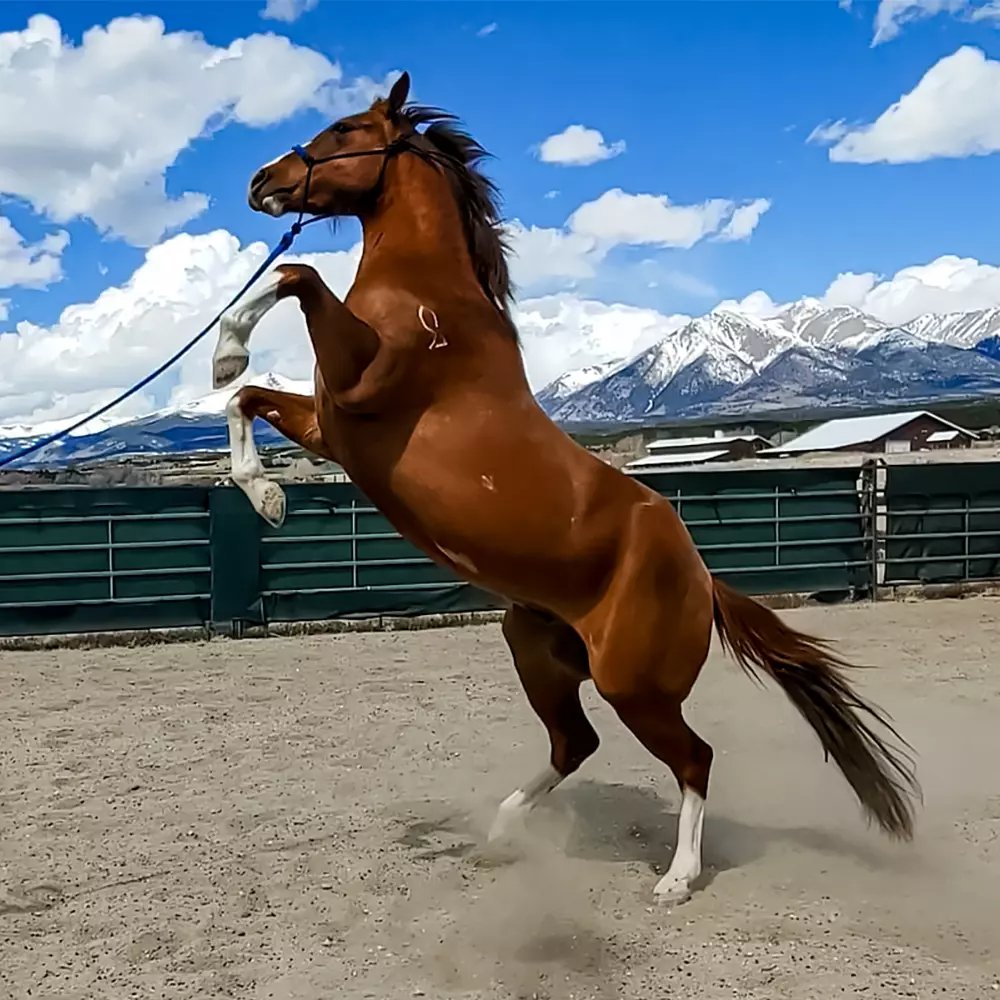}}
	\end{minipage}
	\begin{minipage}[b]{0.170\linewidth}
		\centering
		\includegraphics[width = 0.99\linewidth]{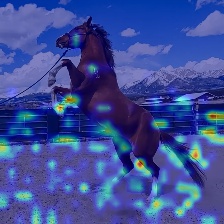}
	\end{minipage}
	\begin{minipage}[b]{0.170\linewidth}
		\centering
		\includegraphics[width = 0.99\linewidth]{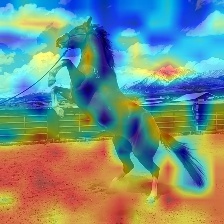}	
	\end{minipage}
	\begin{minipage}[b]{0.170\linewidth}
		\centering
		\includegraphics[width = 0.99\linewidth]{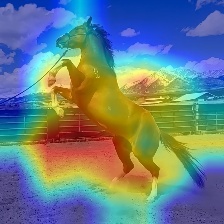}
	\end{minipage}
	\begin{minipage}[b]{0.170\linewidth}
		\centering
		\includegraphics[width = 0.99\linewidth]{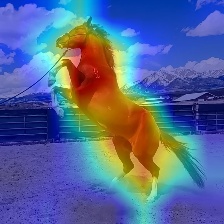}	
	\end{minipage}
\vskip 0.04in

    \begin{minipage}[b]{0.09\linewidth}
		\centering
		\textbf{\texttt{\fontsize{7.5pt}{\baselineskip}\selectfont (a-2)}}
	\end{minipage}
    \begin{minipage}[b]{0.170\linewidth}
		\centering
		\fcolorbox{green}{yellow}{\includegraphics[width = 0.99\linewidth]{CAM_Visualization/Example_Images/horse.jpg}}
	\end{minipage}
	\begin{minipage}[b]{0.170\linewidth}
		\centering
		\includegraphics[width = 0.99\linewidth]{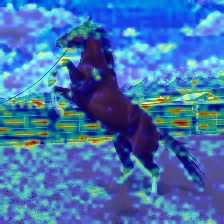}
	\end{minipage}
	\begin{minipage}[b]{0.170\linewidth}
		\centering
		\includegraphics[width = 0.99\linewidth]{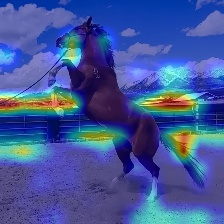}	
	\end{minipage}
	\begin{minipage}[b]{0.170\linewidth}
		\centering
		\includegraphics[width = 0.99\linewidth]{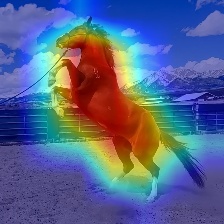}	
	\end{minipage}
	\begin{minipage}[b]{0.170\linewidth}
		\centering
		\includegraphics[width = 0.99\linewidth]{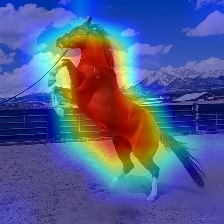}	
	\end{minipage}
\vskip 0.04in

    \begin{minipage}[b]{0.09\linewidth}
		\centering
		\textbf{\texttt{\fontsize{7.5pt}{\baselineskip}\selectfont (b-1)}}
	\end{minipage}
	\begin{minipage}[b]{0.170\linewidth}
		\centering
        \fcolorbox{green}{yellow}{\includegraphics[width = 0.99\linewidth]{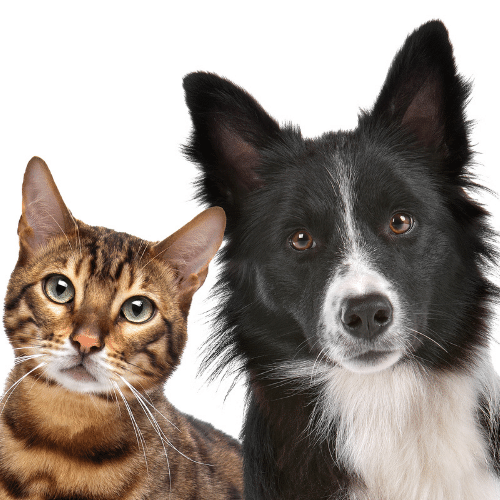}}
	\end{minipage}
	\begin{minipage}[b]{0.170\linewidth}
		\centering
		\includegraphics[width = 0.99\linewidth]{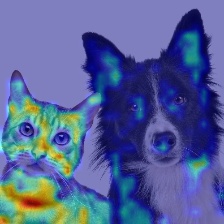}
	\end{minipage}
	\begin{minipage}[b]{0.170\linewidth}
		\centering
		\includegraphics[width = 0.99\linewidth]{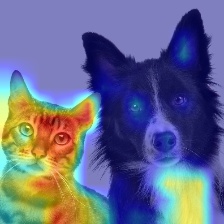}	
	\end{minipage}
	\begin{minipage}[b]{0.170\linewidth}
		\centering
		\includegraphics[width = 0.99\linewidth]{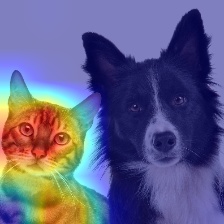}
	\end{minipage}
	\begin{minipage}[b]{0.170\linewidth}
		\centering
		\includegraphics[width = 0.99\linewidth]{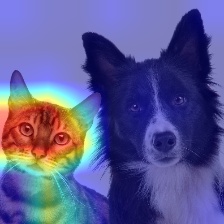}	
	\end{minipage}
\vskip 0.04in

    \begin{minipage}[b]{0.09\linewidth}
		\centering
		\textbf{\texttt{\fontsize{7.5pt}{\baselineskip}\selectfont (b-2)}}
	\end{minipage}
    \begin{minipage}[b]{0.170\linewidth}
		\centering
		\fcolorbox{green}{yellow}{\includegraphics[width = 0.99\linewidth]{CAM_Visualization/Example_Images/cat_and_dog.png}}
	\end{minipage}
	\begin{minipage}[b]{0.170\linewidth}
		\centering
		\includegraphics[width = 0.99\linewidth]{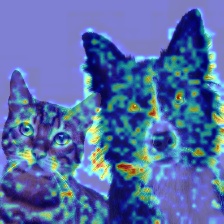}
	\end{minipage}
	\begin{minipage}[b]{0.170\linewidth}
		\centering
		\includegraphics[width = 0.99\linewidth]{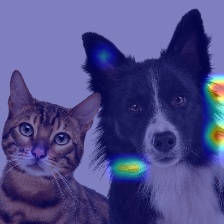}	
	\end{minipage}
	\begin{minipage}[b]{0.170\linewidth}
		\centering
		\includegraphics[width = 0.99\linewidth]{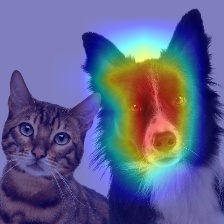}	
	\end{minipage}
	\begin{minipage}[b]{0.170\linewidth}
		\centering
		\includegraphics[width = 0.99\linewidth]{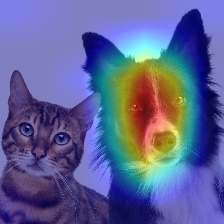}	
	\end{minipage}

\end{center}
\vskip -0.2in
	\caption{Attention map visualization for 
	Swin Transformer ImageNet-1K pretrained models. \textbf{(a-1)} and \textbf{(b-1)} Swin-V1-Tiny [2, 2, 6, 2]. \textbf{(a-2)} and \textbf{(b-2)} Swin-V1-Base [2, 2, 18, 2]. Numbers in square brackets indicate how many Swin Transformer blocks in each stage. We choose the output of last Swin Transformer block in each stage, to generate the CAM visualization results.}
	\label{Fig.inspiration_knowledge_distillation}
\vskip -0.15in
\end{figure}

To improve the efficiency of feature-based KD, we will not mimic each feature map 
in the teacher model. \textit{Instead, we find the critical feature maps from the teacher model as learning targets.} 
We use the tiny- and base-sized Swin Transformer~\cite{liu2021swin} pretrained models as an example to apply the Class Activation Map~\cite{selvaraju2017grad} (CAM) for feature map visualization~\cite{wang2020score}, as shown in Figure~\ref{Fig.inspiration_knowledge_distillation}. The Swin Transformer blocks are organized into four stages with different feature map resolutions. We use the outputs of the last Swin Transformer block in each stage as the representatives. By comparing the CAM results in \textbf{(a-1)} and \textbf{(a-2)}, we find the attention is focused on 
local features in the early stages, while focused on 
global features of the target object in the later stages. Moreover, even though the tiny- and base-sized models provide the same classification result for the horse input image, the CAM from early stages (i.e., stages 1 to 3) are quite different. This phenomenon inspires us that \textit{it is more effective to mimic the feature maps from later stages of the vision transformer models.} By comparing the CAM results in \textbf{(b-1)} and \textbf{(b-2)}, the tiny-sized model classifies the input as an Egyptian cat, and the base-sized model classifies it as a Border collie. Different classified labels influence the CAM to pay attention to totally different features of a cat and a collie, respectively. \textit{It inspires us to enable mimic feature learning only when the teacher and student models have the same classification labels; otherwise, skip the mimic behavior.}

Denoting distillation losses for the hard label, soft logits and feature maps are $\textit{\textbf{L}}_{hard\_label}^{prune}$, $\textit{\textbf{L}}_{soft\_logits}^{prune}$, $\textit{\textbf{L}}_{feature}^{prune}$, respectively, and their weight factors are: $\alpha, \beta, \gamma$, then the overall sparse pruning loss $\textit{\textbf{L}}_{prune}$ is calculated as follows:
\begin{equation}
\small
\textit{\textbf{L}}_{prune}=\alpha*\textit{\textbf{L}}_{hard\_label}^{prune}+\beta*\textit{\textbf{L}}_{soft\_logits}^{prune}+\gamma*\textit{\textbf{L}}_{feature}^{prune}
\end{equation}
The 2:4 structured sparse pruning workflow minimizes the $\textit{\textbf{L}}_{prune}$ loss w.r.t weight parameters of $\textbf{\emph{M}}_{SF}$ model.

\setlength{\parskip}{0.6em}
\noindent\textbf{Sparse-distillation-aware QAT} aims to further compress the sparse floating-point model $\textbf{\emph{M}}_{SF}$ as the sparse quantized model $\textbf{\emph{M}}_{SQ}$ on data format, i.e., quantize from the floating-point formats to INT8 or INT4. We mainly discuss the QAT strategy 
for the following reasons. \textit{From the performance perspective}, QAT can achieve the same deployment efficiency with the toolkit
~\cite{nvidiatensorrt}. \textit{From the accuracy perspective}, QAT learns the scale factor adjustment during 
training
, so the learned scale factor leads to less quantization noise and a better accuracy compensation effect. 
Moreover, compression by 2:4 fine-grained structured sparsity needs the \textbf{\textit{premise}}~\cite{mishra2021accelerating} to access the training set and undergo a fine-tuning process. \textit{So we can fully utilize the training set and fine-tuning process to calibrate the quantization scale factor and boost the accuracy of quantized 
model.} 
\setlength{\parskip}{0em}

We borrow the KD idea 
and jointly learn to calibrate the quantization scale factor from the teacher model's hard label prediction, soft logits
, and feature maps from critical layers. Unlike the 
sparse pruning workflow in which 
$\textbf{\emph{M}}_{DF}$ model serves as the teacher and 
$\textbf{\emph{M}}_{SF}$ model serves as the student, in the QAT process, 
$\textbf{\emph{M}}_{SF}$ model serves as the teacher, and 
$\textbf{\emph{M}}_{SQ}$ model serves as the student.\footnote{Using the dense floating-point model serves as the teacher in the QAT process is not recommended, even though it usually has better accuracy than the 2:4 sparse floating-point model. Because based on the previous study~\cite{mirzadeh2020improved}~\cite{yu2021minimally}, the distillation effectiveness will drop if the teacher and student models have a noticeable gap in scale or data format.} \textit{Another difference between the KD strategies in two workflows is a weight factor 
to multiply the feature-based calibration result from each critical layer.} The value of each weight factor is determined by the feature-based distillation loss between the corresponding layers from 
$\textbf{\emph{M}}_{DF}$ and 
$\textbf{\emph{M}}_{SF}$ models.

Usually, after the 2:4 structured sparse pruning workflow, 
$\textbf{\emph{M}}_{DF}$ and 
$\textbf{\emph{M}}_{SF}$ models have similar accuracy. So intuitively, if the distillation loss for the feature map of a specific layer between 
$\textbf{\emph{M}}_{DF}$ and $\textbf{\emph{M}}_{SF}$ models is still significant, \textit{it indicates this layer has little influence on the model's final 
accuracy} and vice versa. So if the distillation loss value is larger, then we give a smaller weight factor for the corresponding feature-based calibration loss, to indicate even the quantization compression leads to the difference between 
$\textbf{\emph{M}}_{SF}$ and $\textbf{\emph{M}}_{SQ}$ models; however, \textit{\textbf{this difference has a low probability of having the real influence on the quantized model's final 
accuracy}}. That's the reason why we named \textbf{GPUSQ-ViT} quantization workflow as \textbf{sparse-distillation-aware QAT}. 
Denoting calibration losses for the hard label, soft logits and feature maps are $\textit{\textbf{L}}_{hard\_label}^{calibrate}$, $\textit{\textbf{L}}_{soft\_logits}^{calibrate}$, $\textit{\textbf{L}}_{feature}^{calibrate}$, respectively, and their weight factors are still: $\alpha, \beta, \gamma$, then the overall quantization calibration loss $\textit{\textbf{L}}_{calibrate}$ is calculated as follows:
\begin{equation}
\small
\textit{\textbf{L}}_{calibrate}=\alpha*\textit{\textbf{L}}_{hard\_label}^{calibrate}+\beta*\textit{\textbf{L}}_{soft\_logits}^{calibrate}+\gamma*\textit{\textbf{L}}_{feature}^{calibrate}
\end{equation}
The sparse-distillation-aware QAT workflow minimizes the $\textit{\textbf{L}}_{calibrate}$ loss w.r.t weight parameters of $\textbf{\emph{M}}_{SQ}$ model. The details about each loss items in \textbf{GPUSQ-ViT} are provided in \textbf{Algorithm 1} 
in \textbf{Appendix}.

\section{Experiments}
\label{sec:experiments}

For the experiments in this paper, we choose PyTorch~\cite{paszke2017automatic} with version 1.12.0 as the framework to implement all algorithms. The results of the dense model training, sparse compression, and QAT experiments are obtained with A100~\cite{nvidiaa100} GPU clusters. The acceleration performance results for deployment are obtained with A100 GPU and AGX Orin chip~\cite{nvidiaorin} to represent the server and edge device scenarios, respectively. Both A100 and Orin have the Tensor Core~\cite{nvidiatc} support for 2:4 structured sparsity and mixed-precision calculation among FP16, INT8, and INT4. All the reference algorithms use the default data type provided in public repositories.

\subsection{Compression efficacy for classification task}
\label{subsection_effectiveness_classification}

\begin{table}[htp]
\vskip -0.1in
\centering
\resizebox{0.995\linewidth}{!}{
\begin{tabular}{l|l|ll|ll|ll}
\toprule
\textbf{Model}                        & \textbf{Method}             & \textbf{Input}            & \textbf{Format} & \textbf{\begin{tabular}[c]{@{}l@{}}Params (M)\end{tabular}} & \textbf{\begin{tabular}[c]{@{}l@{}}FLOPs (G)\end{tabular}} & \textbf{\begin{tabular}[c]{@{}l@{}}Top-1 Acc(\%)\end{tabular}} & \textbf{\begin{tabular}[c]{@{}l@{}}Top-5 Acc(\%)\end{tabular}} \\
\midrule
\multirow{10}{*}{\textbf{DeiT-Tiny}}  & \textit{Baseline}           & \multirow{10}{*}{224$^2$} & FP32            & 5.72                                                          & 1.30                                                         & 72.2                                                             & 91.1                                                             \\
                                      & S$^2$ViTE                   &                           & FP32            & 4.21                                                          & 0.99                                                         & 70.1                                                             & 90.1                                                             \\
                                      & SViTE                       &                           & FP32            & 3.46                                                          & 0.86                                                         & 71.8                                                             & 90.6                                                             \\
                                      & MiniViT                     &                           & FP32            & 3.09                                                          & 1.30                                                         & 72.8                                                             & 91.6                                                             \\
                                      & PS-ViT                      &                           & FP32            & 3.08                                                          & 0.70                                                         & 72.0                                                             & 91.0                                                             \\
                                      & UVC                         &                           & FP32            & 3.08                                                          & 0.69                                                         & 71.8                                                             & 90.6                                                             \\
                                      & FQ-ViT                      &                           & INT8            & 1.43                                                          & 1.27                                                         & 71.6                                                             & 90.6                                                             \\
                                      & \textit{\textbf{GPUSQ-ViT}} &                           & INT8            & 0.90 \color{blue}{(6.4$\times$)}                                            & 0.04 \color{blue}{(31$\times$)}                                            & 72.4 \color{green}{(+0.2)}                                                      & 90.9 \color{red}{(-0.2)}                                                      \\
                                      & Q-ViT                       &                           & INT4            & 0.72                                                          & 0.34                                                         & 71.6                                                             & 90.5                                                             \\
                                      & \textit{\textbf{GPUSQ-ViT}} &                           & INT4            & 0.45 \color{blue}{(12.7$\times$)}                                           & 0.02 \color{blue}{(62$\times$)}                                            & 71.7 \color{red}{(-0.5)}                                                      & 90.6 \color{red}{(-0.5)}                                                      \\ \hline
\multirow{15}{*}{\textbf{DeiT-Small}} & \textit{Baseline}           & \multirow{15}{*}{224$^2$} & FP32            & 22.05                                                         & 4.60                                                         & 79.9                                                             & 95.0                                                             \\
                                      & DyViT                       &                           & FP32            & 26.90                                                         & 3.70                                                         & 82.0                                                             & 95.5                                                             \\
                                      & MultiViT                    &                           & FP32            & 16.76                                                         & 2.90                                                         & 79.9                                                             & 94.9                                                             \\
                                      & IA-RED$^2$                  &                           & FP32            & 14.99                                                         & 3.10                                                         & 79.1                                                             & 94.5                                                             \\
                                      & S$^2$ViTE                   &                           & FP32            & 14.60                                                         & 2.12                                                         & 79.2                                                             & 94.6                                                             \\
                                      & MiniViT                     &                           & FP32            & 11.45                                                         & 4.70                                                         & 80.7                                                             & 95.6                                                             \\
                                      & PS-ViT                      &                           & FP32            & 12.46                                                         & 2.59                                                         & 79.4                                                             & 94.7                                                             \\
                                      & UVC                         &                           & FP32            & 12.70                                                         & 2.65                                                         & 79.4                                                             & 94.7                                                             \\
                                      & SViTE                       &                           & FP32            & 8.90                                                          & 1.38                                                         & 79.4                                                             & 94.7                                                             \\
                                      & PTQ-ViT                     &                           & INT8            & 5.51                                                          & 5.67                                                         & 78.1                                                             & 94.2                                                             \\
                                      & PTQ4ViT                     &                           & INT8            & 5.51                                                          & 3.45                                                         & 79.5                                                             & 94.7                                                             \\
                                      & FQ-ViT                      &                           & INT8            & 5.51                                                          & 4.61                                                         & 79.2                                                             & 94.6                                                             \\
                                      & \textit{\textbf{GPUSQ-ViT}} &                           & INT8            & 3.46 \color{blue}{(6.4$\times$)}                                            & 0.14 \color{blue}{(31$\times$)}                                            & 80.3 \color{green}{(+0.4)}                                                      & 95.1 \color{green}{+0.1)}                                                      \\
                                      & Q-ViT                       &                           & INT4            & 2.76                                                          & 1.22                                                         & 80.1                                                             & 94.9                                                             \\
                                      & \textit{\textbf{GPUSQ-ViT}} &                           & INT4            & 1.73 \color{blue}{(12.7$\times$)}                                           & 0.07 \color{blue}{(62$\times$)}                                            & 79.3 \color{red}{(-0.6)}                                                      & 94.8 \color{red}{(-0.2)}                                                      \\ \hline
\multirow{14}{*}{\textbf{DeiT-Base}}  & \textit{Baseline}           & \multirow{14}{*}{224$^2$} & FP32            & 86.57                                                         & 17.60                                                        & 81.8                                                             & 95.6                                                             \\
                                      & MultiViT                    &                           & FP32            & 64.93                                                         & 11.20                                                        & 82.3                                                             & 96.0                                                             \\
                                      & IA-RED$^2$                  &                           & FP32            & 58.01                                                         & 11.80                                                        & 80.9                                                             & 95.0                                                             \\
                                      & S$^2$ViTE                   &                           & FP32            & 56.80                                                         & 11.77                                                        & 82.2                                                             & 95.8                                                             \\
                                      & MiniViT                     &                           & FP32            & 44.10                                                         & 17.70                                                        & 83.2                                                             & 96.5                                                             \\
                                      & PS-ViT                      &                           & FP32            & 48.22                                                         & 9.80                                                         & 81.5                                                             & 95.4                                                             \\
                                      & UVC                         &                           & FP32            & 39.40                                                         & 8.01                                                         & 80.6                                                             & 94.5                                                             \\
                                      & SViTE                       &                           & FP32            & 34.80                                                         & 7.48                                                         & 81.3                                                             & 95.3                                                             \\
                                      & PTQ-ViT                     &                           & INT8            & 21.64                                                         & 20.10                                                        & 81.3                                                             & 95.2                                                             \\
                                      & FQ-ViT                      &                           & INT8            & 21.64                                                         & 17.48                                                        & 81.2                                                             & 95.2                                                             \\
                                      & PTQ4ViT                     &                           & INT8            & 21.64                                                         & 13.10                                                        & 81.5                                                             & 95.3                                                             \\
                                      & \textit{\textbf{GPUSQ-ViT}} &                           & INT8            & 13.55 \color{blue}{(6.4$\times$)}                                           & 0.55 \color{blue}{(31$\times$)}                                            & 82.9 \color{green}{(+1.1)}                                                      & 96.4 \color{green}{(+0.8)}                                                      \\
                                      & PTQ4ViT                     &                           & INT4            & 10.82                                                         & 6.94                                                         & 75.9                                                             & 95.3                                                             \\
                                      & \textit{\textbf{GPUSQ-ViT}} &                           & INT4            & 6.78 \color{blue}{(12.7$\times$)}                                           & 0.28 \color{blue}{(62$\times$)}                                            & 81.6 \color{red}{(-0.2)}                                                      & 95.5 \color{red}{(-0.1)}                                                      \\ \hline
\multirow{6}{*}{\textbf{DeiT-Base}}   & \textit{Baseline}           & \multirow{6}{*}{384$^2$}  & FP32            & 86.86                                                         & 55.60                                                        & 82.9                                                             & 96.2                                                             \\
                                      & IA-RED                      &                           & FP32            & 54.31                                                         & 34.70                                                        & 81.9                                                             & 95.7                                                             \\
                                      & MiniViT                     &                           & FP32            & 44.39                                                         & 56.90                                                        & 84.7                                                             & 97.2                                                             \\
                                      & PTQ4ViT                     &                           & INT8            & 21.71                                                         & 41.70                                                        & 82.9                                                             & 96.3                                                             \\
                                      & \textit{\textbf{GPUSQ-ViT}} &                           & INT8            & 13.62 \color{blue}{(6.4$\times$)}                                           & 1.74 \color{blue}{(31$\times$)}                                            & 82.9 \color{green}{(+0.0)}                                                      & 96.3 \color{green}{(+0.1)}                                                      \\
                                      & \textit{\textbf{GPUSQ-ViT}} &                           & INT4            & 6.81 \color{blue}{(12.7$\times$)}                                           & 0.87 \color{blue}{(62$\times$)}                                            & 82.4 \color{red}{(-0.5)}                                                      & 96.1 \color{red}{(-0.1)}                                                      \\
\hline
\hline
\multirow{8}{*}{\textbf{Swin-Tiny}}   & \textit{Baseline}           & \multirow{8}{*}{224$^2$}  & FP32            & 28.29                                                         & 4.49                                                         & 81.2                                                             & 95.5                                                             \\
                                      & Dyn-ViT                     &                           & FP32            & 19.80                                                         & 4.00                                                         & 80.9                                                             & 95.4                                                             \\
                                      & MiniViT                     &                           & FP32            & 12.00                                                         & 4.60                                                         & 81.4                                                             & 95.7                                                             \\
                                      & FQ-ViT                      &                           & INT8            & 7.07                                                          & 4.39                                                         & 80.5                                                             & 95.2                                                             \\
                                      & PTQ4ViT                     &                           & INT8            & 7.07                                                          & 3.37                                                         & 81.2                                                             & 95.4                                                             \\
                                      & \textit{\textbf{GPUSQ-ViT}} &                           & INT8            & 4.43 \color{blue}{(6.4$\times$)}                                            & 0.14 \color{blue}{(31$\times$)}                                            & 81.2 \color{green}{(+0.0)}                                                      & 95.5 \color{green}{(+0.0)}                                                      \\
                                      & Q-ViT                       &                           & INT4            & 3.54                                                          & 1.10                                                         & 80.6                                                             & 95.2                                                             \\
                                      & \textit{\textbf{GPUSQ-ViT}} &                           & INT4            & 2.21 \color{blue}{(12.7$\times$)}                                           & 0.07 \color{blue}{(62$\times$)}                                            & 80.7 \color{red}{(-0.5)}                                                      & 95.3 \color{red}{(-0.2)}                                                      \\ \hline
\multirow{7}{*}{\textbf{Swin-Small}}  & \textit{Baseline}           & \multirow{7}{*}{224$^2$}  & FP32            & 49.61                                                         & 8.75                                                         & 83.2                                                             & 96.2                                                             \\
                                      & Dyn-ViT                     &                           & FP32            & 34.73                                                         & 6.90                                                         & 83.2                                                             & 96.3                                                             \\
                                      & MiniViT                     &                           & FP32            & 26.46                                                         & 8.93                                                         & 83.6                                                             & 97.0                                                             \\
                                      & FQ-ViT                      &                           & INT8            & 12.40                                                         & 8.77                                                         & 82.7                                                             & 96.1                                                             \\
                                      & PTQ4ViT                     &                           & INT8            & 12.40                                                         & 6.56                                                         & 83.1                                                             & 96.2                                                             \\
                                      & \textit{\textbf{GPUSQ-ViT}} &                           & INT8            & 7.77 \color{blue}{(6.4$\times$)}                                            & 0.27 \color{blue}{(31$\times$)}                                            & 83.1 \color{red}{(-0.1)}                                                      & 96.3 \color{green}{(+0.1)}                                                      \\
                                      & \textit{\textbf{GPUSQ-ViT}} &                           & INT4            & 3.88 \color{blue}{(12.7$\times$)}                                           & 0.14 \color{blue}{(62$\times$)}                                            & 82.8 \color{red}{(-0.4)}                                                      & 96.2 \color{green}{(+0.0)}                                                      \\ \hline
\multirow{7}{*}{\textbf{Swin-Base}}   & \textit{Baseline}           & \multirow{7}{*}{224$^2$}  & FP32            & 87.77                                                         & 15.44                                                        & 83.5                                                             & 96.5                                                             \\
                                      & Dyn-ViT                     &                           & FP32            & 61.44                                                         & 12.10                                                        & 83.4                                                             & 96.4                                                             \\
                                      & MiniViT                     &                           & FP32            & 46.44                                                         & 15.71                                                        & 84.3                                                             & 97.3                                                             \\
                                      & FQ-ViT                      &                           & INT8            & 21.94                                                         & 15.33                                                        & 83.0                                                             & 96.3                                                             \\
                                      & PTQ4ViT                     &                           & INT8            & 21.94                                                         & 11.58                                                        & 83.2                                                             & 96.3                                                             \\
                                      & \textit{\textbf{GPUSQ-ViT}} &                           & INT8            & 13.73 \color{blue}{(6.4$\times$)}                                           & 0.48 \color{blue}{(31$\times$)}                                            & 83.4 \color{red}{(-0.1)}                                                      & 96.4 \color{red}{(-0.1)}                                                      \\
                                      & \textit{\textbf{GPUSQ-ViT}} &                           & INT4            & 6.87 \color{blue}{(12.7$\times$)}                                           & 0.24 \color{blue}{(62$\times$)}                                            & 83.2 \color{red}{(-0.3)}                                                      & 96.3 \color{red}{(-0.2)}                                                      \\ \hline
\multirow{5}{*}{\textbf{Swin-Base}}   & \textit{Baseline}           & \multirow{5}{*}{384$^2$}  & FP32            & 87.90                                                         & 47.11                                                        & 84.5                                                             & 97.0                                                             \\
                                      & MiniViT                     &                           & FP32            & 47.00                                                         & 49.40                                                        & 85.5                                                             & 97.6                                                             \\
                                      & PTQ4ViT                     &                           & INT8            & 21.98                                                         & 35.33                                                        & 84.3                                                             & 96.8                                                             \\
                                      & \textit{\textbf{GPUSQ-ViT}} &                           & INT8            & 13.77 \color{blue}{(6.4$\times$)}                                           & 1.47 \color{blue}{(31$\times$)}                                            & 84.4 \color{red}{(-0.1)}                                                      & 97.0 \color{green}{(0.0)}                                                       \\
                                      & \textit{\textbf{GPUSQ-ViT}} &                           & INT4            & 6.88 \color{blue}{(12.7$\times$)}                                           & 0.74 \color{blue}{(62$\times$)}                                            & 84.4 \color{red}{(-0.1)}                                                      & 96.9 \color{red}{(-0.1)}                                                      \\
\bottomrule
\end{tabular}
}
\vskip -0.1in
\caption{Compare the 
model size and FLOPs of \textbf{GPUSQ-ViT} with state-of-the-art compression methods on classification task.}
\label{Table.classification_sota_compare}
\vskip -0.25in
\end{table}

To evaluate the compression efficacy of \textbf{GPUSQ-ViT} and make the comparison with prior arts on the image classification task, DeiT~\cite{touvron2021training}\footnote{\tiny{\url{https://github.com/facebookresearch/deit}}\label{deit}} and Swin Transformer~\cite{liu2021swin}\footnote{\tiny{\url{https://github.com/microsoft/Swin-Transformer}}\label{swin_transformer}} are chosen as the experiment target models. 
For the state-of-the-art vision transformer compression methods, we choose the Dyn-ViT~\cite{rao2021dynamicvit}, MiniViT~\cite{zhang2022minivit}, UVC~\cite{yu2022unified}, PS-ViT~\cite{tang2022patch}, IA-RED$^2$~\cite{pan2021ia}, MultiViT~\cite{hou2022multi}, SViTE~\cite{chen2021chasing} and S$^2$ViTE~\cite{chen2021chasing} as the reference methods from sparse pruning category, and we choose the FQ-ViT~\cite{lin2022fq}, Q-ViT~\cite{li2022q}, PTQ-ViT~\cite{liu2021post} and PTQ4ViT~\cite{yuan2021ptq4vit} as the reference methods from quantization category. For \textbf{GPUSQ-ViT}, the loss adjustment factors for hard label, soft logits and feature-based losses apply $\alpha=1$, $\beta=10$, and $\gamma=5$), respectively. The model size and FLOPs comparison results are shown in Table~\ref{Table.classification_sota_compare}.

\begin{table}[htp]
\vskip -0.1in
\centering
\resizebox{0.955\linewidth}{!}{
\begin{tabular}{l|l|ll|ll|ll}
\toprule
\multirow{2}{*}{\textbf{Model}}      & \multirow{2}{*}{\textbf{Method}} & \multirow{2}{*}{\textbf{Input}} & \multirow{2}{*}{\textbf{Format}} & \multicolumn{2}{l|}{\textbf{NVIDIA A100 GPU}}                                                                                   & \multicolumn{2}{l}{\textbf{NVIDIA AGX Orin}}                                                                                   \\ \cline{5-8} 
                                     &                                  &                                 &                                  & \textbf{\begin{tabular}[c]{@{}l@{}}FPS\\ (BS=1)\end{tabular}} & \textbf{\begin{tabular}[c]{@{}l@{}}FPS\\ (BS=256)\end{tabular}} & \textbf{\begin{tabular}[c]{@{}l@{}}FPS\\ (BS=1)\end{tabular}} & \textbf{\begin{tabular}[c]{@{}l@{}}FPS\\ (BS=64)\end{tabular}} \\
\midrule
\multirow{3}{*}{\textbf{DeiT-Tiny}}  & \textit{Baseline}                & \multirow{3}{*}{224$^2$}        & FP32                             & 3067                                                          & 14934                                                           & 2671                                                          & 4005                                                           \\
                                     & \textit{\textbf{GPUSQ-ViT}}      &                                 & INT8                             & 3864 \color{blue}{(1.26$\times$)}                                           & 38978 \color{blue}{(2.60$\times$)}                                            & 3232 \color{blue}{(1.21$\times$)}                                           & 7329 \color{blue}{(1.83$\times$)}                                            \\
                                     & \textit{\textbf{GPUSQ-ViT}}      &                                 & INT4                             & 4263 \color{blue}{(1.39$\times$)}                                           & 51224 \color{blue}{(3.43$\times$)}                                            & 4193 \color{blue}{(1.57$\times$)}                                           & 8531 \color{blue}{(2.13$\times$)}                                            \\ \hline
\multirow{3}{*}{\textbf{DeiT-Small}} & \textit{Baseline}                & \multirow{3}{*}{224$^2$}        & FP32                             & 1256                                                          & 5277                                                            & 877                                                           & 1280                                                           \\
                                     & \textit{\textbf{GPUSQ-ViT}}      &                                 & INT8                             & 1629 \color{blue}{(1.30$\times$)}                                           & 13359 \color{blue}{(2.53$\times$)}                                            & 1096 \color{blue}{(1.25$\times$)}                                           & 2291 \color{blue}{(1.79$\times$)}                                            \\
                                     & \textit{\textbf{GPUSQ-ViT}}      &                                 & INT4                             & 1809 \color{blue}{(1.44$\times$)}                                           & 17775 \color{blue}{(3.37$\times$)}                                            & 1447 \color{blue}{(1.65$\times$)}                                           & 2701 \color{blue}{(2.11$\times$)}                                            \\ \hline
\multirow{3}{*}{\textbf{DeiT-Base}}  & \textit{Baseline}                & \multirow{3}{*}{224$^2$}        & FP32                             & 485                                                           & 1682                                                            & 351                                                           & 513                                                            \\
                                     & \textit{\textbf{GPUSQ-ViT}}      &                                 & INT8                             & 645 \color{blue}{(1.33$\times$)}                                            & 4136 \color{blue}{(2.46$\times$)}                                             & 453 \color{blue}{(1.29$\times$)}                                            & 939 \color{blue}{(1.83$\times$)}                                             \\
                                     & \textit{\textbf{GPUSQ-ViT}}      &                                 & INT4                             & 714 \color{blue}{(1.47$\times$)}                                            & 5643 \color{blue}{(3.35$\times$)}                                             & 569 \color{blue}{(1.62$\times$)}                                            & 1206 \color{blue}{(2.35$\times$)}                                            \\ \hline
\multirow{3}{*}{\textbf{DeiT-Base}}  & \textit{Baseline}                & \multirow{3}{*}{384$^2$}        & FP32                             & 256                                                           & 689                                                             & 233                                                           & 303                                                            \\
                                     & \textit{\textbf{GPUSQ-ViT}}      &                                 & INT8                             & 350 \color{blue}{(1.37$\times$)}                                            & 1730 \color{blue}{(2.51$\times$)}                                             & 308 \color{blue}{(1.32$\times$)}                                            & 561 \color{blue}{(1.85$\times$)}                                             \\
                                     & \textit{\textbf{GPUSQ-ViT}}      &                                 & INT4                             & 394 \color{blue}{(1.54$\times$)}                                            & 2315 \color{blue}{(3.36$\times$)}                                             & 371 \color{blue}{(1.59$\times$)}                                            & 761 \color{blue}{(2.51$\times$)}                                             \\
\hline
\hline
\multirow{3}{*}{\textbf{Swin-Tiny}}  & \textit{Baseline}                & \multirow{3}{*}{224$^2$}        & FP32                             & 621                                                           & 2907                                                            & 544                                                           & 968                                                            \\
                                     & \textit{\textbf{GPUSQ-ViT}}      &                                 & INT8                             & 807 \color{blue}{(1.30$\times$)}                                            & 6975 \color{blue}{(2.40$\times$)}                                             & 675 \color{blue}{(1.24$\times$)}                                            & 1946 \color{blue}{(2.01$\times$)}                                            \\
                                     & \textit{\textbf{GPUSQ-ViT}}      &                                 & INT4                             & 910 \color{blue}{(1.46$\times$)}                                            & 9911 \color{blue}{(3.41$\times$)}                                             & 892 \color{blue}{(1.64$\times$)}                                            & 2275 \color{blue}{(2.35$\times$)}                                            \\ \hline
\multirow{3}{*}{\textbf{Swin-Small}} & \textit{Baseline}                & \multirow{3}{*}{224$^2$}        & FP32                             & 330                                                           & 1802                                                            & 309                                                           & 631                                                            \\
                                     & \textit{\textbf{GPUSQ-ViT}}      &                                 & INT8                             & 426 \color{blue}{(1.29$\times$)}                                            & 4411 \color{blue}{(2.45$\times$)}                                             & 392 \color{blue}{(1.27$\times$)}                                            & 1306 \color{blue}{(2.07$\times$)}                                            \\
                                     & \textit{\textbf{GPUSQ-ViT}}      &                                 & INT4                             & 510 \color{blue}{(1.55$\times$)}                                            & 5942 \color{blue}{(3.30$\times$)}                                             & 516 \color{blue}{(1.67$\times$)}                                            & 1521 \color{blue}{(2.41$\times$)}                                            \\ \hline
\multirow{3}{*}{\textbf{Swin-Base}}  & \textit{Baseline}                & \multirow{3}{*}{224$^2$}        & FP32                             & 282                                                           & 1261                                                            & 247                                                           & 433                                                            \\
                                     & \textit{\textbf{GPUSQ-ViT}}      &                                 & INT8                             & 388 \color{blue}{(1.37$\times$)}                                            & 3226 \color{blue}{(2.56$\times$)}                                             & 309 \color{blue}{(1.25$\times$)}                                            & 842 \color{blue}{(1.94$\times$)}                                             \\
                                     & \textit{\textbf{GPUSQ-ViT}}      &                                 & INT4                             & 485 \color{blue}{(1.72$\times$)}                                            & 4071 \color{blue}{(3.22$\times$)}                                             & 410 \color{blue}{(1.66$\times$)}                                            & 1063 \color{blue}{(2.45$\times$)}                                            \\ \hline
\multirow{3}{*}{\textbf{Swin-Base}}  & \textit{Baseline}                & \multirow{3}{*}{384$^2$}        & FP32                             & 154                                                           & 531                                                             & 140                                                           & 226                                                            \\
                                     & \textit{\textbf{GPUSQ-ViT}}      &                                 & INT8                             & 226 \color{blue}{(1.47$\times$)}                                            & 1310 \color{blue}{(2.47$\times$)}                                             & 180 \color{blue}{(1.28$\times$)}                                            & 414 \color{blue}{(1.83$\times$)}                                             \\
                                     & \textit{\textbf{GPUSQ-ViT}}      &                                 & INT4                             & 369 \color{blue}{(1.79$\times$)}                                            & 1747 \color{blue}{(3.29$\times$)}                                             & 238 \color{blue}{(1.69$\times$)}                                            & 562 \color{blue}{(2.48$\times$)}                                             \\
\bottomrule
\end{tabular}
}
\vskip -0.1in
\caption{Deployment efficiency of \textbf{GPUSQ-ViT} compressed DeiT and Swin Transformer models on NVIDIA GPUs. The latency is measured with batch size 1 on a single A100 GPU and AGX Orin. The throughput is measured with batch size fixed to 256 on a single A100 GPU and with batch size fixed to 64 on a single AGX Orin.}
\label{Table.classification_fps}
\vskip -0.1in
\end{table}

\fboxsep = 0.0pt    
\fboxrule = 1.0pt   
\begin{figure}[htb]
\begin{center}
    \centering
    \begin{minipage}[b]{0.130\linewidth}
		\centering
		\textbf{\texttt{\fontsize{5.5pt}{\baselineskip}\selectfont }}
	\end{minipage}
    \begin{minipage}[b]{0.420\linewidth}
		\centering
		\textbf{\texttt{\fontsize{5.5pt}{\baselineskip}\selectfont |\qquad Swin Transformer Tiny\qquad|}}
	\end{minipage}
	\begin{minipage}[b]{0.420\linewidth}
		\centering
		\textbf{\texttt{\fontsize{5.5pt}{\baselineskip}\selectfont |\qquad Swin Transformer Base\qquad|}}
	\end{minipage}
\vskip -0.07in

    \begin{minipage}[b]{0.130\linewidth}
		\centering
		\textbf{\texttt{\fontsize{5.5pt}{\baselineskip}\selectfont Input}}
	\end{minipage}
    \begin{minipage}[b]{0.130\linewidth}
		\centering
		\textbf{\texttt{\fontsize{5.5pt}{\baselineskip}\selectfont Baseline}}
	\end{minipage}
	\begin{minipage}[b]{0.130\linewidth}
		\centering
		\textbf{\texttt{\fontsize{5.5pt}{\baselineskip}\selectfont INT8}}
	\end{minipage}
	\begin{minipage}[b]{0.130\linewidth}
		\centering
		\textbf{\texttt{\fontsize{5.5pt}{\baselineskip}\selectfont INT4}}
	\end{minipage}
	\begin{minipage}[b]{0.130\linewidth}
		\centering
		\textbf{\texttt{\fontsize{5.5pt}{\baselineskip}\selectfont Baseline}}
	\end{minipage}
	\begin{minipage}[b]{0.130\linewidth}
		\centering
		\textbf{\texttt{\fontsize{5.5pt}{\baselineskip}\selectfont INT8}}
	\end{minipage}
	\begin{minipage}[b]{0.130\linewidth}
		\centering
		\textbf{\texttt{\fontsize{5.5pt}{\baselineskip}\selectfont INT4}}
	\end{minipage}
\vskip 0.00in

	\begin{minipage}[b]{0.130\linewidth}
		\centering
        \fcolorbox{green}{yellow}{\includegraphics[width = 0.99\linewidth]{CAM_Visualization/Example_Images/horse.jpg}}
	\end{minipage}
	\begin{minipage}[b]{0.130\linewidth}
		\centering
		\includegraphics[width = 0.99\linewidth]{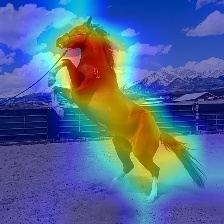}
	\end{minipage}
	\begin{minipage}[b]{0.130\linewidth}
		\centering
		\includegraphics[width = 0.99\linewidth]{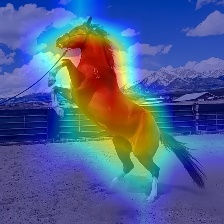}	
	\end{minipage}
	\begin{minipage}[b]{0.130\linewidth}
		\centering
		\includegraphics[width = 0.99\linewidth]{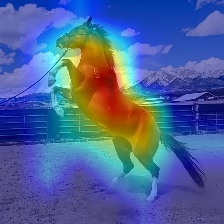}
	\end{minipage}
	\begin{minipage}[b]{0.130\linewidth}
		\centering
		\includegraphics[width = 0.99\linewidth]{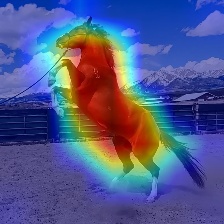}	
	\end{minipage}
	\begin{minipage}[b]{0.130\linewidth}
		\centering
		\includegraphics[width = 0.99\linewidth]{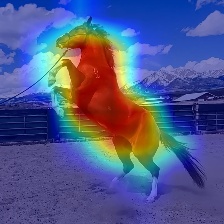}	
	\end{minipage}
	\begin{minipage}[b]{0.130\linewidth}
		\centering
		\includegraphics[width = 0.99\linewidth]{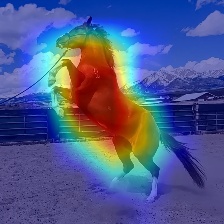}
	\end{minipage}
\vskip 0.00in

	\begin{minipage}[b]{0.130\linewidth}
		\centering
        \fcolorbox{green}{yellow}{\includegraphics[width = 0.99\linewidth]{CAM_Visualization/Example_Images/cat_and_dog.png}}
	\end{minipage}
	\begin{minipage}[b]{0.130\linewidth}
		\centering
		\includegraphics[width = 0.99\linewidth]{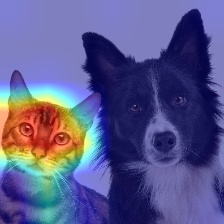}
	\end{minipage}
	\begin{minipage}[b]{0.130\linewidth}
		\centering
		\includegraphics[width = 0.99\linewidth]{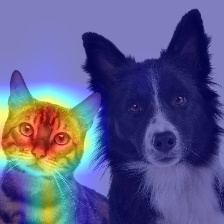}	
	\end{minipage}
	\begin{minipage}[b]{0.130\linewidth}
		\centering
		\includegraphics[width = 0.99\linewidth]{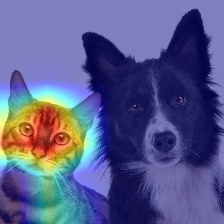}
	\end{minipage}
	\begin{minipage}[b]{0.130\linewidth}
		\centering
		\includegraphics[width = 0.99\linewidth]{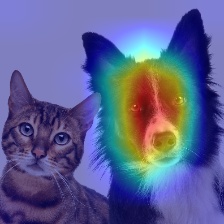}	
	\end{minipage}
	\begin{minipage}[b]{0.130\linewidth}
		\centering
		\includegraphics[width = 0.99\linewidth]{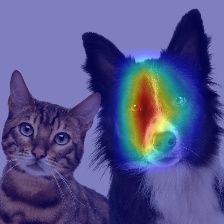}	
	\end{minipage}
	\begin{minipage}[b]{0.130\linewidth}
		\centering
		\includegraphics[width = 0.99\linewidth]{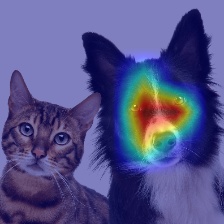}
	\end{minipage}
\vskip 0.00in

	\begin{minipage}[b]{0.130\linewidth}
		\centering
        \fcolorbox{green}{yellow}{\includegraphics[width = 0.99\linewidth]{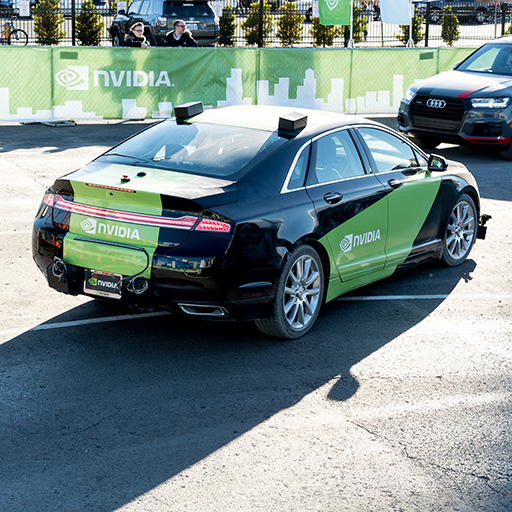}}
	\end{minipage}
	\begin{minipage}[b]{0.130\linewidth}
		\centering
		\includegraphics[width = 0.99\linewidth]{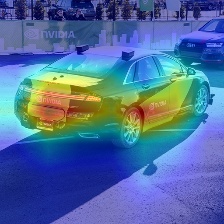}
	\end{minipage}
	\begin{minipage}[b]{0.130\linewidth}
		\centering
		\includegraphics[width = 0.99\linewidth]{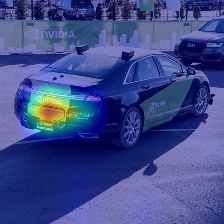}	
	\end{minipage}
	\begin{minipage}[b]{0.130\linewidth}
		\centering
		\includegraphics[width = 0.99\linewidth]{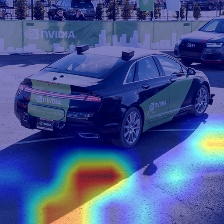}
	\end{minipage}
	\begin{minipage}[b]{0.130\linewidth}
		\centering
		\includegraphics[width = 0.99\linewidth]{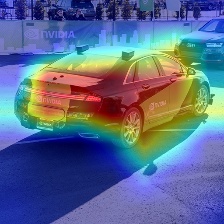}	
	\end{minipage}
	\begin{minipage}[b]{0.130\linewidth}
		\centering
		\includegraphics[width = 0.99\linewidth]{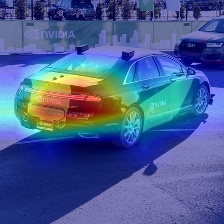}	
	\end{minipage}
	\begin{minipage}[b]{0.130\linewidth}
		\centering
		\includegraphics[width = 0.99\linewidth]{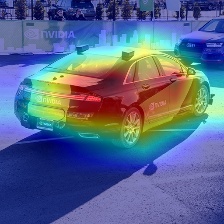}
	\end{minipage}
\vskip 0.00in


\end{center}
\vskip -0.2in
	\caption{
	CAM visualization for Swin Transformer baseline dense models and \textbf{GPUSQ-ViT} compressed INT8 and INT4 models.}
	\label{Fig.cam_between_dense_and_compressed}
\vskip -0.25in
\end{figure}

We can apply \textbf{GPUSQ-ViT} to compress each vision 
model as INT8 and INT4 versions. For INT8 compressed models, \textbf{GPUSQ-ViT} can bring 6.4$\times$ reduction for model size and 31$\times$ reduction for FLOPs with negligible accuracy drop. For INT4 compressed models, \textbf{GPUSQ-ViT} can get 12.7$\times$ and 62$\times$ reduction for model size and FLOPs with a small accuracy drop. Compared with both sparse pruning and quantization prior arts, \textbf{GPUSQ-ViT} can steadily provide more reduction for model size and FLOPs.

Moreover, \textbf{GPUSQ-ViT} can greatly boost the compressed models' deployment efficiency on GPUs with TensorRT toolkit~\cite{nvidiatensorrt} support of 2:4 sparsity. For INT8 compressed models, \textbf{GPUSQ-ViT} can bring 
1.26-1.47$\times$ and 2.4-2.6$\times$ improvement for various DeiT and Swin Transformer models of latency and throughput on A100 GPU, and 
1.21-1.32$\times$ and 1.79-2.07$\times$ improvement of latency and throughput on AGX Orin. For INT4 compressed models, \textbf{GPUSQ-ViT} can bring 
1.39-1.79$\times$ and 3.22-3.43$\times$ improvement of latency and throughput on A100 GPU, and 
1.57-1.69$\times$ and 2.11-2.51$\times$ improvement of latency and throughput on AGX Orin, as shown in Table~\ref{Table.classification_fps}. 

To compare between dense and \textbf{GPUSQ-ViT} compressed models in visualization, we apply CAM for tiny- and base-sized Swin Transformer models' attention on final norm layer. The results are shown in Figure~\ref{Fig.cam_between_dense_and_compressed}.

\subsection{Compression efficacy for detection task}
\label{subsection_effectiveness_detection}

To evaluate the compression efficacy of \textbf{GPUSQ-ViT} on the object detection task, Mask R-CNN~\cite{he2017mask}\footnote{\tiny{\url{https://github.com/SwinTransformer/Swin-Transformer-Object-Detection}}\label{swin_detection}}, DETR~\cite{carion2020end}\footnote{\tiny{\url{https://github.com/facebookresearch/detr}}\label{detr}} and 
Deformable-DETR~\cite{zhu2020deformable} \footnote{\tiny{\url{https://github.com/fundamentalvision/Deformable-DETR}}\label{deformable_detr}} are chosen as the 
target models. 
\textbf{GPUSQ-ViT} compression results on COCO dataset ~\cite{lin2014microsoft} are shown in Table~\ref{Table.object_detection}.

\begin{table}[!htb]
\vskip -0.1in
\resizebox{0.985\linewidth}{!}{
\begin{tabular}{l|l|ll|ll|ll}
\toprule
\multirow{2}{*}{\textbf{Model}}                                                        & \multirow{2}{*}{\textbf{Backbone}} & \multirow{2}{*}{\textbf{Method}} & \multirow{2}{*}{\textbf{Format}} & \multirow{2}{*}{\textbf{Params (M)}} & \multirow{2}{*}{\textbf{FLOPs (G)}} & \multirow{2}{*}{\textbf{bbox mAP}} & \multirow{2}{*}{\textbf{segm mAP}} \\
                                                                                       &                                    &                                  &                                  &                                      &                                     &                                    &                                    \\
\midrule
\multirow{6}{*}{\textbf{Mask R-CNN}}                                                   & \multirow{3}{*}{Swin-Tiny}         & \textit{Baseline}                & FP32                             & 48                                   & 267                                 & 46.0                               & 41.6                               \\
                                                                                       &                                    & \textit{\textbf{GPUSQ-ViT}}      & INT8                             & 7.5 \color{blue}{(6.4$\times$)}                    & 8.8 \color{blue}{(30.5$\times$)}                  & 46.0 \color{green}{(+0.0)}                        & 41.6 \color{green}{(+0.0)}                        \\
                                                                                       &                                    & \textit{\textbf{GPUSQ-ViT}}      & INT4                             & 3.8 \color{blue}{(12.7$\times$)}                   & 4.4 \color{blue}{(61.0$\times$)}                  & 45.7 \color{red}{(-0.3)}                        & 41.4 \color{red}{(-0.2)}                        \\ \cline{2-8} 
                                                                                       & \multirow{3}{*}{Swin-Small}        & \textit{Baseline}                & FP32                             & 69                                   & 359                                 & 48.5                               & 43.3                               \\
                                                                                       &                                    & \textit{\textbf{GPUSQ-ViT}}      & INT8                             & 10.8 \color{blue}{(6.4$\times$)}                   & 11.8 \color{blue}{(30.5$\times$)}                 & 48.6 \color{green}{(+0.1)}                        & 43.4 \color{green}{(+0.1)}                        \\
                                                                                       &                                    & \textit{\textbf{GPUSQ-ViT}}      & INT4                             & 5.4 \color{blue}{(12.7$\times$)}                   & 5.9 \color{blue}{(61.0$\times$)}                  & 48.3 \color{red}{(-0.2)}                        & 43.2 \color{red}{(-0.1)}                        \\ \hline
\multirow{9}{*}{\textbf{\begin{tabular}[c]{@{}l@{}}Cascade\\ Mask R-CNN\end{tabular}}} & \multirow{3}{*}{Swin-Tiny}         & \textit{Baseline}                & FP32                             & 86                                   & 745                                 & 48.1                               & 41.7                               \\
                                                                                       &                                    & \textit{\textbf{GPUSQ-ViT}}      & INT8                             & 13.4 \color{blue}{(6.4$\times$)}                   & 24.4 \color{blue}{(30.5$\times$)}                 & 48.1 (+0.0)                        & 41.8 \color{green}{(+0.1)}                        \\
                                                                                       &                                    & \textit{\textbf{GPUSQ-ViT}}      & INT4                             & 6.8 \color{blue}{(12.7$\times$)}                   & 12.2 \color{blue}{(61.0$\times$)}                 & 47.8 \color{red}{(-0.3)}                        & 41.5 \color{red}{(-0.2)}                        \\ \cline{2-8} 
                                                                                       & \multirow{3}{*}{Swin-Small}        & \textit{Baseline}                & FP32                             & 107                                  & 838                                 & 51.9                               & 45.0                               \\
                                                                                       &                                    & \textit{\textbf{GPUSQ-ViT}}      & INT8                             & 16.7 \color{blue}{(6.4$\times$)}                   & 27.5 \color{blue}{(30.5$\times$)}                 & 52.0 \color{green}{(+0.1)}                        & 45.2 \color{green}{(+0.2)}                        \\
                                                                                       &                                    & \textit{\textbf{GPUSQ-ViT}}      & INT4                             & 8.4 \color{blue}{(12.7$\times$)}                   & 13.7 \color{blue}{(61.0$\times$)}                 & 51.7 \color{red}{(-0.2)}                        & 44.9 \color{red}{(-0.1)}                        \\ \cline{2-8} 
                                                                                       & \multirow{3}{*}{Swin-Base}         & \textit{Baseline}                & FP32                             & 145                                  & 982                                 & 51.9                               & 45.0                               \\
                                                                                       &                                    & \textit{\textbf{GPUSQ-ViT}}      & \textit{\textbf{INT8}}           & 22.7 \color{blue}{(6.4$\times$)}                   & 32.2 \color{blue}{(30.5$\times$)}                 & 52.1 \color{green}{(+0.2)}                        & 45.3 \color{green}{(+0.3)}                        \\
                                                                                       &                                    & \textit{\textbf{GPUSQ-ViT}}      & \textit{\textbf{INT4}}           & 11.4 \color{blue}{(12.7$\times$)}                  & 16.1 \color{blue}{(61.0$\times$)}                 & 51.8 \color{red}{(-0.1)}                        & 44.9 \color{red}{(-0.1)}                        \\
\hline
\hline
\multirow{3}{*}{\textbf{DETR}}                                                         & \multirow{3}{*}{ResNet50}          & \textit{Baseline}                & FP32                             & 41                                   & 86                                  & 42.0                               & N/A                                \\
                                                                                       &                                    & \textit{\textbf{GPUSQ-ViT}}      & INT8                             & 6.4 \color{blue}{(6.4$\times$)}                    & 2.8 \color{blue}{(30.5$\times$)}                  & 42.0 \color{green}{(+0.0)}                        & N/A                                \\
                                                                                       &                                    & \textit{\textbf{GPUSQ-ViT}}      & INT4                             & 3.2 \color{blue}{(12.7$\times$)}                   & 1.4 \color{blue}{(61.0$\times$)}                  & 41.7 \color{red}{(-0.3)}                        & N/A                                \\ \hline
\multirow{3}{*}{\textbf{\begin{tabular}[c]{@{}l@{}}Deformable\\ DETR\end{tabular}}}    & \multirow{3}{*}{ResNet50}          & \textit{Baseline}                & FP32                             & 40                                   & 173                                 & 44.5                               & N/A                                \\
                                                                                       &                                    & \textit{\textbf{GPUSQ-ViT}}      & INT8                             & 6.3 \color{blue}{(6.4$\times$)}                    & 5.7 \color{blue}{(30.5$\times$)}                  & 44.5 \color{green}{(+0.0)}                        & N/A                                \\
                                                                                       &                                    & \textit{\textbf{GPUSQ-ViT}}      & INT4                             & 3.1 \color{blue}{(12.7$\times$)}                   & 2.8 \color{blue}{(61.0$\times$)}                  & 44.1 \color{red}{(-0.4)}                        & N/A                                \\
\bottomrule
\end{tabular}
}
\vskip -0.1in
\caption{Effectiveness of \textbf{GPUSQ-ViT} on object detection task.}
\label{Table.object_detection}
\vskip -0.2in
\end{table}

\subsection{Compression efficacy for segmentation task}
\label{subsection_effectiveness_segmentation}

To evaluate the compression efficacy of \textbf{GPUSQ-ViT} on the semantic segmentation
 task, UPerNet~\cite{xiao2018unified}\footnote{\tiny{\url{https://github.com/SwinTransformer/Swin-Transformer-Semantic-Segmentation}}\label{swin_segmentation}} is chosen as the 
target model. 
\textbf{GPUSQ-ViT} compression results on ADE20K dataset~\cite{zhou2019semantic} are shown in Table~\ref{Table.semantic_segmentation}. 

\begin{table}[!htp]
\vskip -0.1in
\resizebox{0.985\linewidth}{!}{
\begin{tabular}{l|l|ll|ll|ll}
\toprule
\multirow{2}{*}{\textbf{Model}}   & \multirow{2}{*}{\textbf{Backbone}} & \multirow{2}{*}{\textbf{Method}} & \multirow{2}{*}{\textbf{Format}} & \multirow{2}{*}{\textbf{Params (M)}} & \multirow{2}{*}{\textbf{FLOPs (G)}} & \multirow{2}{*}{\textbf{Mean IoU (\%)}} & \multirow{2}{*}{\textbf{Pixel Acc. (\%)}} \\
                                  &                                    &                                  &                                  &                                      &                                     &                                         &                                           \\
\midrule
\multirow{9}{*}{\textbf{UPerNet}} & \multirow{3}{*}{Swin-Tiny}         & \textit{Baseline}                & FP32                             & 60                                   & 945                                 & 44.51                                   & 81.09                                     \\
                                  &                                    & \textit{\textbf{GPUSQ-ViT}}      & INT8                             & 9.4 \color{blue}{(6.4$\times$)}                    & 31.2 \color{blue}{(30.3$\times$)}                 & 44.47 \color{red}{(-0.04)}                           & 81.01 \color{red}{(-0.08)}                             \\
                                  &                                    & \textit{\textbf{GPUSQ-ViT}}      & INT4                             & 4.7 \color{blue}{(12.7$\times$)}                   & 15.6 \color{blue}{(60.6$\times$)}                 & 43.93 \color{red}{(-0.58)}                           & 80.89 \color{red}{(-0.20)}                             \\ \cline{2-8} 
                                  & \multirow{3}{*}{Swin-Small}        & \textit{Baseline}                & FP32                             & 81                                   & 1038                                & 47.64                                   & 82.45                                     \\
                                  &                                    & \textit{\textbf{GPUSQ-ViT}}      & INT8                             & 12.7 \color{blue}{(6.4$\times$)}                   & 34.3 \color{blue}{(30.3$\times$)}                 & 47.66 \color{green}{(+0.02)}                           & 82.41 \color{red}{(-0.04)}                             \\
                                  &                                    & \textit{\textbf{GPUSQ-ViT}}      & INT4                             & 6.4 \color{blue}{(12.7$\times$)}                   & 17.1 \color{blue}{(60.6$\times$)}                 & 47.15 \color{red}{(-0.49)}                           & 82.30 \color{red}{(-0.15)}                             \\ \cline{2-8} 
                                  & \multirow{3}{*}{Swin-Base}         & \textit{Baseline}                & FP32                             & 121                                  & 1188                                & 48.13                                   & 82.37                                     \\
                                  &                                    & \textit{\textbf{GPUSQ-ViT}}      & \textit{\textbf{INT8}}           & 18.9 \color{blue}{(6.4$\times$)}                   & 39.2 \color{blue}{(30.3$\times$)}                 & 48.18 \color{green}{(+0.05)}                           & 82.43 \color{green}{(+0.06)}                             \\
                                  &                                    & \textit{\textbf{GPUSQ-ViT}}      & \textit{\textbf{INT4}}           & 9.5 \color{blue}{(12.7$\times$)}                   & 19.6 \color{blue}{(60.6$\times$)}                 & 47.86 \color{red}{(-0.27)}                           & 82.19 \color{red}{(-0.18)}                             \\
\bottomrule
\end{tabular}
}
\vskip -0.1in
\caption{Effectiveness of \textbf{GPUSQ-ViT} on semantic segmentation.}
\label{Table.semantic_segmentation}
\vskip -0.15in
\end{table}

\textbf{GPUSQ-ViT} provides good compression effects on detection and segmentation tasks in Table~\ref{Table.object_detection} and~\ref{Table.semantic_segmentation} with small accuracy gap to the dense baseline models.

\subsection{GPUSQ-ViT with unsupervised learning}
\label{subsection_unsupervised_learning}

Because the compressed model can learn the representation of target from dense model's prediction when lacking ground-truth label annotations, so \textbf{GPUSQ-ViT} can still work well in 
unsupervised training
, as shown in Table~\ref{Table.unsupervised_learning_styles}. 

\begin{table}[!htp]
\vskip -0.1in
\resizebox{0.995\linewidth}{!}{
\begin{tabular}{l|l|ll|ll}
\toprule
\multirow{2}{*}{\textbf{Model}} & \multirow{2}{*}{\textbf{Input}} & \multicolumn{2}{l|}{\textbf{GPUSQ-ViT (INT8)}}                                                                                      & \multicolumn{2}{l}{\textbf{GPUSQ-ViT (INT4)}}                                                                                       \\ \cline{3-6} 
                                &                                 & \textbf{\begin{tabular}[c]{@{}l@{}}Top-1\\ Acc(\%)\end{tabular}} & \textbf{\begin{tabular}[c]{@{}l@{}}Top-5\\ Acc(\%)\end{tabular}} & \textbf{\begin{tabular}[c]{@{}l@{}}Top-1\\ Acc(\%)\end{tabular}} & \textbf{\begin{tabular}[c]{@{}l@{}}Top-5\\ Acc(\%)\end{tabular}} \\
\midrule
\textbf{DeiT-Tiny}              & 224$^2$                         & 72.0 \color{red}{(-0.2)}                                                      & 90.8 \color{red}{(-0.3)}                                                      & 71.4 \color{red}{(-0.8)}                                                      & 90.2 \color{red}{(-0.9)}                                                      \\ \hline
\textbf{DeiT-Small}             & 224$^2$                         & 79.8 \color{red}{(-0.1)}                                                      & 94.9 \color{red}{(-0.1)}                                                      & 79.2 \color{red}{(-0.7)}                                                      & 94.2 \color{red}{(-0.8)}                                                      \\ \hline
\textbf{DeiT-Base}              & 224$^2$                         & 82.0 \color{green}{(+0.2)}                                                      & 95.7 \color{green}{(+0.1)}                                                      & 81.1 \color{red}{(-0.7)}                                                      & 95.0 \color{red}{(-0.6)}                                                      \\ \hline
\textbf{DeiT-Base}              & 384$^2$                         & 82.5 \color{red}{(-0.4)}                                                      & 95.9 \color{red}{(-0.3)}                                                      & 82.0 \color{red}{(-0.9)}                                                      & 95.7 \color{red}{(-0.5)}                                                      \\ \hline
\textbf{Swin-Tiny}              & 224$^2$                         & 80.8 \color{red}{(-0.4)}                                                      & 95.2 \color{red}{(-0.3)}                                                      & 80.3 \color{red}{(-0.9)}                                                      & 94.9 \color{red}{(-0.6)}                                                      \\ \hline
\textbf{Swin-Small}             & 224$^2$                         & 82.7 \color{red}{(-0.5)}                                                      & 95.9 \color{red}{(-0.3)}                                                      & 82.3 \color{red}{(-0.9)}                                                      & 95.7 \color{red}{(-0.5)}                                                      \\ \hline
\textbf{Swin-Base}              & 224$^2$                         & 82.9 \color{red}{(-0.6)}                                                      & 96.1 \color{red}{(-0.4)}                                                      & 82.5 \color{red}{(-1.0)}                                                      & 95.7 \color{red}{(-0.8)}                                                      \\ \hline
\textbf{Swin-Base}              & 384$^2$                         & 83.9 \color{red}{(-0.6)}                                                      & 96.6 \color{red}{(-0.4)}                                                      & 83.7 \color{red}{(-0.8)}                                                      & 96.4 \color{red}{(-0.6)}                                                      \\
\bottomrule
\end{tabular}
}
\vskip -0.1in
\caption{Effectiveness of \textbf{GPUSQ-ViT} in unsupervised learning.}
\label{Table.unsupervised_learning_styles}
\vskip -0.1in
\end{table}

\subsection{Ablation study of \textbf{GPUSQ-ViT}}
\label{subsection_ablation_study}

\begin{table}[!htp]
\resizebox{0.995\linewidth}{!}{
\begin{tabular}{l|llll|ll|ll}
\toprule
\multirow{3}{*}{\textbf{Model}}                                                         & \multirow{3}{*}{\textbf{\begin{tabular}[c]{@{}l@{}}Factor $\alpha$\end{tabular}}} & \multirow{3}{*}{\textbf{\begin{tabular}[c]{@{}l@{}}Factor $\beta$\end{tabular}}} & \multirow{3}{*}{\textbf{\begin{tabular}[c]{@{}l@{}}Factor $\gamma$\end{tabular}}} & \multirow{3}{*}{\textbf{\begin{tabular}[c]{@{}l@{}}Enable QAT\\ Weight Factor\end{tabular}}} & \multicolumn{2}{l|}{\textbf{GPUSQ-ViT (INT8)}}                                                                                      & \multicolumn{2}{l}{\textbf{GPUSQ-ViT (INT4)}}                                                                                       \\ \cline{6-9} 
                                                                                        &                                                                                                &                                                                                                &                                                                                                   &                                                                                              & \textbf{\begin{tabular}[c]{@{}l@{}}Top-1\\ Acc(\%)\end{tabular}} & \textbf{\begin{tabular}[c]{@{}l@{}}Top-5\\ Acc(\%)\end{tabular}} & \textbf{\begin{tabular}[c]{@{}l@{}}Top-1\\ Acc(\%)\end{tabular}} & \textbf{\begin{tabular}[c]{@{}l@{}}Top-5\\ Acc(\%)\end{tabular}} \\ 
\midrule
\multirow{8}{*}{\textbf{\begin{tabular}[c]{@{}l@{}}DeiT-Base\\ (224$^2$)\end{tabular}}} & 1                                                                                              & 10                                                                                             & 5                                                                                                 & \Checkmark                                                                    & 82.9 \color{green}{(+1.1)}                                                      & 96.4 \color{green}{(+0.8)}                                                      & 81.6 \color{red}{(-0.2)}                                                      & 95.5 \color{red}{(-0.1)}                                                      \\
                                                                                        & 1                                                                                              & 10                                                                                             & 5                                                                                                 & \XSolidBrush                                                                  & 82.4 \color{green}{(+0.6)}                                                      & 96.1 \color{green}{(+0.5)}                                                      & 80.1 \color{red}{(-1.7)}                                                      & 94.3 \color{red}{(-1.3)}                                                      \\
                                                                                        & 1                                                                                              & 0                                                                                              & 5                                                                                                 & \Checkmark                                                                    & 82.7 \color{green}{(+0.9)}                                                      & 96.2 \color{green}{(+0.6)}                                                      & 81.3 \color{red}{(-0.5)}                                                      & 95.2 \color{red}{(-0.4)}                                                      \\
                                                                                        & 1                                                                                              & 10                                                                                             & 0                                                                                                 & \Checkmark                                                                    & 82.2 \color{green}{(+0.4)}                                                      & 95.8 \color{green}{(+0.2)}                                                      & 80.8 \color{red}{(-1.0)}                                                      & 94.8 \color{red}{(-0.8)}                                                      \\
                                                                                        & 1                                                                                              & 20                                                                                             & 5                                                                                                 & \Checkmark                                                                    & 82.9 \color{green}{(+1.1)}                                                      & 96.4 \color{green}{(+0.8)}                                                      & 81.6 \color{red}{(-0.2)}                                                      & 95.6 \color{green}{(+0.0)}                                                      \\
                                                                                        & 1                                                                                              & 30                                                                                             & 5                                                                                                 & \Checkmark                                                                    & 82.9 \color{green}{(+1.1)}                                                      & 96.5 \color{green}{(+0.9)}                                                      & 81.6 \color{red}{(-0.2)}                                                      & 95.6 \color{green}{(+0.0)}                                                      \\
                                                                                        & 1                                                                                              & 10                                                                                             & 10                                                                                                & \Checkmark                                                                    & 82.8 \color{green}{(+1.0)}                                                      & 96.5 \color{green}{(+0.9)}                                                      & 81.5 \color{red}{(-0.3)}                                                      & 95.5 \color{red}{(-0.1)}                                                      \\
                                                                                        & 1                                                                                              & 10                                                                                             & 2.5                                                                                                & \Checkmark                                                                    & 82.8 \color{green}{(+1.0)}                                                      & 96.5 \color{green}{(+0.9)}                                                      & 81.5 \color{red}{(-0.3)}                                                      & 95.6 \color{green}{(+0.0)}                                                      \\
\hline
\hline
\multirow{8}{*}{\textbf{\begin{tabular}[c]{@{}l@{}}Swin-Base\\ (224$^2$)\end{tabular}}} & 1                                                                                              & 10                                                                                             & 5                                                                                                 & \Checkmark                                                                    & 83.4 \color{red}{(-0.1)}                                                      & 96.4 \color{red}{(-0.1)}                                                      & 83.2 \color{red}{(-0.3)}                                                      & 96.3 \color{red}{(-0.2)}                                                      \\
                                                                                        & 1                                                                                              & 10                                                                                             & 5                                                                                                 & \XSolidBrush                                                                  & 82.9 \color{red}{(-0.6)}                                                      & 96.0 \color{red}{(-0.5)}                                                      & 81.5 \color{red}{(-2.0)}                                                      & 94.9 \color{red}{(-1.6)}                                                      \\
                                                                                        & 1                                                                                              & 0                                                                                              & 5                                                                                                 & \Checkmark                                                                    & 83.2 \color{red}{(-0.3)}                                                      & 96.2 \color{red}{(-0.3)}                                                      & 82.9 \color{red}{(-0.6)}                                                      & 96.0 \color{red}{(-0.5)}                                                      \\
                                                                                        & 1                                                                                              & 10                                                                                             & 0                                                                                                 & \Checkmark                                                                    & 82.7 \color{red}{(-0.8)}                                                      & 95.7 \color{red}{(-0.8)}                                                      & 82.4 \color{red}{(-1.1)}                                                      & 95.5 \color{red}{(-1.0)}                                                      \\
                                                                                        & 1                                                                                              & 20                                                                                             & 5                                                                                                 & \Checkmark                                                                    & 83.4 \color{red}{(-0.1)}                                                      & 96.4 \color{red}{(-0.1)}                                                      & 83.2 \color{red}{(-0.3)}                                                      & 96.3 \color{red}{(-0.2)}                                                      \\
                                                                                        & 1                                                                                              & 30                                                                                             & 5                                                                                                 & \Checkmark                                                                    & 83.4 \color{red}{(-0.1)}                                                      & 96.4 \color{red}{(-0.1)}                                                      & 83.2 \color{red}{(-0.3)}                                                      & 96.4 \color{red}{(-0.1)}                                                      \\
                                                                                        & 1                                                                                              & 10                                                                                             & 10                                                                                                & \Checkmark                                                                    & 83.3 \color{red}{(-0.2)}                                                      & 96.4 \color{red}{(-0.1)}                                                      & 83.1 \color{red}{(-0.4)}                                                      & 96.3 \color{red}{(-0.2)}                                                      \\
                                                                                        & 1                                                                                              & 10                                                                                             & 2.5                                                                                                & \Checkmark                                                                    & 83.3 \color{red}{(-0.2)}                                                      & 96.4 \color{red}{(-0.1)}                                                      & 83.1 \color{red}{(-0.4)}                                                      & 96.4 \color{red}{(-0.1)}                                                      \\
\bottomrule
\end{tabular}
}
\vskip -0.1in
\caption{Ablation study of the loss adjustment factors and sparse-distillation-aware weight factors of \textbf{GPUSQ-ViT} method.}
\label{Table.ablation}
\vskip -0.2in
\end{table}

The ablation study to measure the influence of the different adjustment factors for the hard label, soft logits, and feature-based losses ($\alpha$, $\beta$, $\gamma$) and enabling sparse-distillation-aware weight factor on \textbf{GPUSQ-ViT} compressed model accuracy is shown in Table~\ref{Table.ablation}. From the ablation results, we can find enabling sparse-distillation-aware weight factor has an apparent boost for the compressed models' accuracy. Such a boost effect is more influential on INT4 than INT8 model, because disabling this weight factor will see a more significant drop in INT4 compressed model. The potential reason is sparse-distillation-aware weight factor indicates how much influence the quantization error from each critical layer has on the final accuracy. So the distillation process can focus on mimicking the layers with more accuracy influence, which is more effective for limited quantized bits.
Then we can find disabling the feature-based distillation will lead to a more severe influence than disabling the soft logits distillation. It indicates that mimicking feature maps is very helpful for accuracy compensation in \textbf{GPUSQ-ViT} compression. Finally, we can find \textbf{GPUSQ-ViT} is relatively robust to the soft logits and feature-based loss adjustment factors, i.e., within the close range of $\beta=10$ and $\gamma=5$ the accuracy of compressed models are stable.

\section{Conclusion and limitation}
\label{sec:conclusion}

This paper is inspired by GPU's acceleration characteristic for 2:4 sparse pattern with various low-precision operators to design the \textbf{GPUSQ-ViT} compression method, which can boost deployment efficiency for various 
vision transformer models of benchmarking tasks on NVIDIA GPUs.

We should notice a potential \textit{limitation}. If the structured sparse support is changed or extended to support other 
patterns like 1:4 or 2:16, \textbf{GPUSQ-ViT} needs to make the according adjustments to fit the new or more sparse patterns.

\vskip 0.1in
\noindent \textbf{Acknowledgements} \quad\quad This work is supported by National Natural Science Foundation of China (No. 62071127, U1909207 and 62101137), Shanghai Municipal Science and Technology Major Project (No.2021SHZDZX0103), Shanghai Natural Science Foundation (No. 23ZR1402900), Zhejiang Lab Project (No. 2021KH0AB05).


{\small
\bibliographystyle{ieee_fullname}
\bibliography{egbib}
}

\end{document}